\documentclass{article}
\usepackage{iclr2020_conference}

\usepackage[utf8]{inputenc} %
\usepackage[T1]{fontenc}    %
\usepackage[pagebackref=true,breaklinks=true,colorlinks,citecolor={gray}, bookmarks=false]{hyperref}      %
\usepackage{url}            %
\usepackage{booktabs}       %
\usepackage{amsfonts}       %
\usepackage{nicefrac}       %
\usepackage{microtype}      %
\usepackage{multirow}
\usepackage{graphicx}
\usepackage{makecell}
\usepackage{amsmath}
\usepackage{subfigure}
\usepackage{bbm}
\usepackage{verbatim}
\usepackage{times}

\iclrfinalcopy

\usepackage{epsfig}
\usepackage{graphicx}

\usepackage{adjustbox}
\usepackage{array}
\usepackage{booktabs}
\usepackage{colortbl}
\usepackage{float,wrapfig}
\usepackage{hhline}
\usepackage{multirow}

\usepackage{amsmath,amsfonts,amsthm,amssymb}
\usepackage{bm}
\usepackage{nicefrac}
\usepackage{microtype}

\usepackage{changepage}
\usepackage{extramarks}
\usepackage{fancyhdr}
\usepackage{lastpage}
\usepackage{setspace}
\usepackage{soul}
\usepackage{xspace}

\newcommand*{\mathcolor}{}
\def\mathcolor#1#{\mathcoloraux{#1}}
\newcommand*{\mathcoloraux}[3]{%
  \protect\leavevmode
  \begingroup
    \color#1{#2}#3%
  \endgroup
}

\DeclareMathOperator*{\argmax}{arg\,max}

\newcolumntype{L}[1]{>{\raggedright\let\newline\\\arraybackslash\hspace{0pt}}m{#1}}
\newcolumntype{C}[1]{>{\centering\let\newline\\\arraybackslash\hspace{0pt}}m{#1}}
\newcolumntype{R}[1]{>{\raggedleft\let\newline\\\arraybackslash\hspace{0pt}}m{#1}}

\newcommand{\ignore}[1]{}

\makeatletter
\DeclareRobustCommand\onedot{\futurelet\@let@token\@onedot}
\def\@onedot{\ifx\@let@token.\else.\null\fi\xspace}

\makeatother

\definecolor{MyDarkBlue}{rgb}{0,0.08,1}
\definecolor{MyDarkGreen}{rgb}{0.02,0.6,0.02}
\definecolor{MyDarkRed}{rgb}{0.8,0.02,0.02}
\definecolor{MyDarkOrange}{rgb}{0.40,0.2,0.02}
\definecolor{MyPurple}{RGB}{111,0,255}
\definecolor{MyRed}{rgb}{1.0,0.0,0.0}
\definecolor{MyGold}{rgb}{0.75,0.6,0.12}
\definecolor{MyDarkgray}{rgb}{0.66, 0.66, 0.66}

\newcommand{\myparagraph}[1]{\vspace{-10pt}\paragraph{#1}}

\title{Contrastive Representation Distillation}

\author{%
Yonglong Tian\\
MIT CSAIL\\
\texttt{yonglong@mit.edu}
\And
Dilip Krishnan\\
Google Research\\
\texttt{dilipkay@google.com}
\And
Phillip Isola\\
MIT CSAIL\\
\texttt{phillipi@mit.edu}
}

\begin{document}

\maketitle

\begin{abstract}
  Often we wish to transfer representational knowledge from one neural network to another. Examples include distilling a large network into a smaller one, transferring knowledge from one sensory modality to a second, or ensembling a collection of models into a single estimator. Knowledge distillation, the standard approach to these problems, minimizes the KL divergence between the probabilistic outputs of a teacher and student network. We demonstrate that this objective ignores important \emph{structural} knowledge of the teacher network. This motivates an alternative objective by which we train a student to capture significantly more \emph{information} in the teacher's representation of the data. We formulate this objective as contrastive learning. Experiments demonstrate that our resulting new objective outperforms knowledge distillation and other cutting-edge distillers on a variety of knowledge transfer tasks, including single model compression, ensemble distillation, and cross-modal transfer. Our method sets a new state-of-the-art in many transfer tasks, and sometimes even outperforms the \emph{teacher} network when combined with knowledge distillation. \footnotetext{Code: \url{http://github.com/HobbitLong/RepDistiller}.}
\end{abstract}
\section{Introduction}
\vspace{-5pt}
Knowledge distillation (KD) transfers knowledge from one deep learning model (the teacher) to another (the student). The objective originally proposed by \cite{hinton2015distilling} minimizes the KL divergence between the teacher and student outputs. This formulation makes intuitive sense when the output is a distribution, e.g., a probability mass function over classes. However, often we instead wish to transfer knowledge about a \emph{representation}. For example, in the problem of ``cross-modal distillation", we may wish to transfer the %
representation of an image processing network to a sound \citep{aytar2016soundnet} or to depth \citep{crossmodaldistillation} processing network, such that deep 
features for an image and the associated sound or depth features are highly correlated. In such cases, the KL divergence is undefined.

Representational knowledge is \emph{structured} -- the dimensions exhibit complex interdependencies. The original KD objective introduced in \citep{hinton2015distilling} treats all dimensions as independent, conditioned on the input. Let $\mathbf{y}^T$ be the output of the teacher and $\mathbf{y}^S$ be the output of the student. Then the original KD objective function, $\psi$, has the fully factored form: $\psi(\mathbf{y}^S, \mathbf{y}^T) = \sum_i \phi_i(\mathbf{y}^S_i, \mathbf{y}^T_i)$\footnote{In particular, in \citep{hinton2015distilling}, $\phi_i(a,b) = - a\log b \quad \forall i$}. Such a factored objective is insufficient for transferring structural knowledge, i.e. dependencies between output dimensions $i$ and $j$. This is similar to the situation in image generation where an $L_2$ objective produces blurry results, due to independence assumptions between output dimensions.
    
To overcome this problem, we would like an objective that captures correlations and higher order output dependencies. To achieve this, in this paper we leverage the family of \emph{contrastive} objectives \citep{gutmann2010noise, oord2018representation, arora2019theoretical, hjelm2018learning}. These objective functions have been used successfully in recent years for density estimation and representation learning, especially in self-supervised settings. Here we adapt them to the task of knowledge distillation from one deep network to another. We show that it is important to work in representation space, similar to recent works such as \cite{zagoruyko2016paying, romero2014fitnets}. However, note that the loss functions used in those works do not explicitly try to capture correlations or higher-order dependencies in representational space. 

Our objective maximizes a lower-bound to the mutual information between the teacher and student representations. 
We find that this results in better performance on several knowledge transfer tasks. We conjecture that this is because the contrastive objective better transfers all the information in the teacher's representation, rather than only transferring knowledge about conditionally independent output class probabilities. Somewhat surprisingly, the contrastive objective even improves results on the originally proposed task of distilling knowledge about class probabilities, for example, compressing a large CIFAR100 network into a smaller one that performs almost as well. We believe this is because the correlations between different class probabilities contains useful information that regularizes the learning problem. Our paper forges a connection between two literatures that have evolved mostly independently: \emph{knowledge distillation} and \emph{representation learning}. This connection allows us to leverage strong methods from representation learning to significantly improve the SOTA on knowledge distillation. Our contributions are:

\begin{enumerate}
    \item A contrastive-based objective for transferring knowledge between deep networks.
    \item Applications to model compression, cross-modal transfer, and ensemble distillation.
    \item Benchmarking 12 recent distillation methods; CRD outperforms all other methods, e.g., 57\% average relative improvement over the original KD \citep{hinton2015distilling} \footnote{Average relative improvement = $\frac{1}{N} \sum_{i=1}^{N} \frac{Acc_{crd}^i - Acc_{kd}^i}{Acc_{kd}^i - Acc_{van}^i}$, where $Acc_{crd}^i$, $Acc_{kd}^i$, and $Acc_{van}^i$ represent the accuracies of CRD, KD, and vanilla training of the $i$-th student model, respectively.}, which, surprisingly, performs the second best.
\end{enumerate}

\begin{figure}[t]
\centering
\vspace{-10pt}
\includegraphics[width=\linewidth]{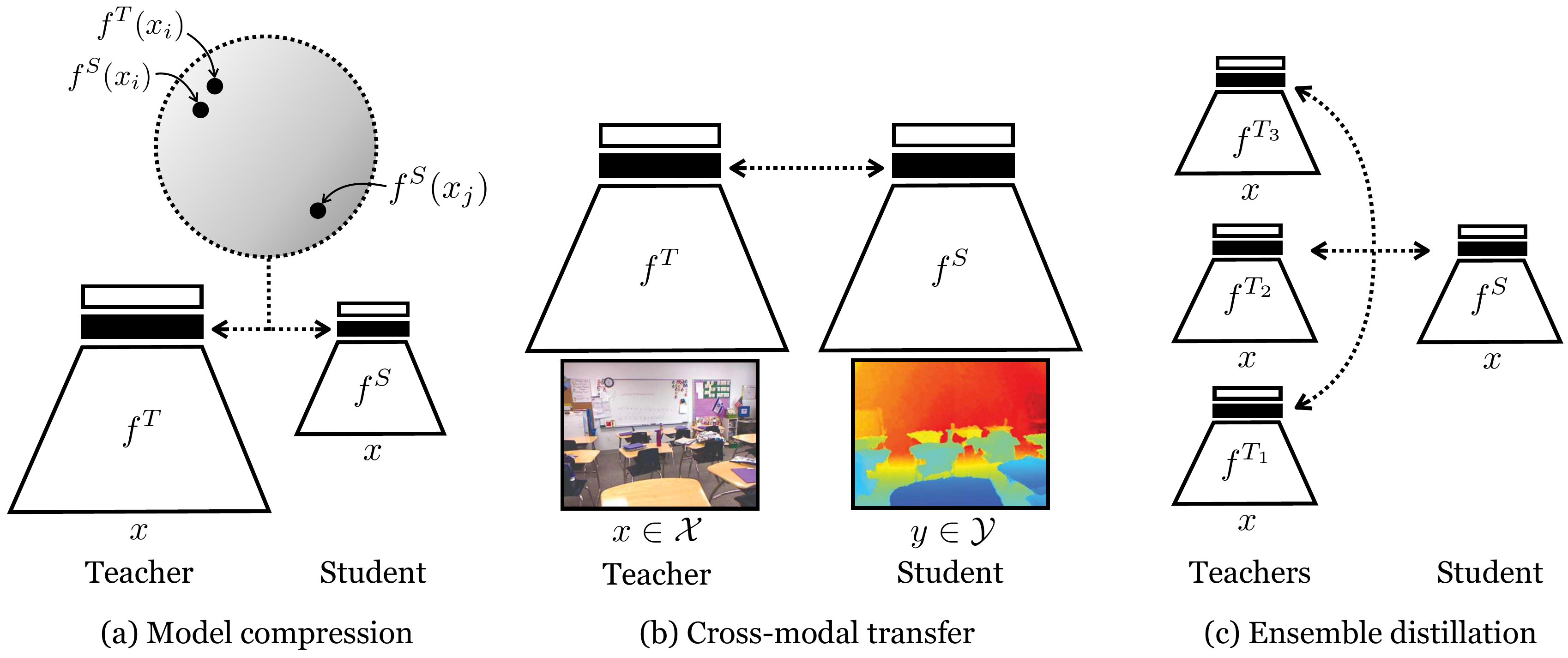}
\vspace{-15pt}
\caption{\small{The three distillation settings we consider: (a) compressing a model, (b) transferring knowledge from one modality (e.g., RGB) to another (e.g., depth), (c) distilling an ensemble of nets into a single network. The constrastive objective encourages the teacher and student to map the same input to close representations (in some metric space), and different inputs to distant representations, as indicated in the shaded circle.}}
\label{fig:methods_diagram}
\vspace{-15pt}
\end{figure}
\section{Related work}
\vspace{-5pt}
The seminal work of \cite{bucilua2006model} and \cite{hinton2015distilling} introduced the idea of knowledge distillation between large, cumbersome models into smaller, faster models without losing too much generalization power. The general motivation was that at training time, the availability of computation allows ``slop" in model size, and potentially faster learning. But computation and memory constraints at inference time necessitate the use of smaller models. \cite{bucilua2006model} achieve this by matching output logits; \cite{hinton2015distilling} introduced the idea of temperature in the softmax outputs to better represent smaller probabilities in the output of a single sample. These smaller probabilities provide useful information about the learned representation of the teacher model; some tradeoff between large temperatures (which increase entropy) or small temperatures tend to provide the highest transfer of knowledge between student and teacher. The method in \citep{li2014learning} was also closely related to \citep{hinton2015distilling}.

Attention transfer \citep{zagoruyko2016paying} focuses on the features maps of the network as opposed to the output logits. Here the idea is to elicit similar response patterns in the teacher and student feature maps (called ``attention"). However, only feature maps with the same spatial resolution can be combined in this approach, which is a significant limitation since it requires student and teacher networks with very similar architectures. This technique achieves state of the art results for distillation (as measured by the generalization of the student network). FitNets \citep{romero2014fitnets} also deal with intermediate representations by using regressions to guide the feature activations of the student network. Since \cite{zagoruyko2016paying} do a weighted form of this regression, they tend to perform better. Other papers \citep{yim2017gift, huang2017like, kim2018paraphrasing, yim2017gift, huang2017like, ahn2019variational, koratana2019lit} have enforced various criteria based on representations. The contrastive objective we use in this paper is the same as that used in CMC~\citep{tian2019contrastive}. But we derive from a different perspective and give a rigorous proof that our objective is a lower bound on mutual information. Our objective is also related to the InfoNCE and NCE objectives introduced in \citep{oord2018representation, gutmann2010noise}. \cite{oord2018representation} use contrastive learning in the context of self-supervised representations learning. They show that their objective maximizes a lower bound on mutual information. A very related approach is used in \citep{hjelm2018learning}. InfoNCE and NCE are closely related but distinct from adversarial learning \citep{goodfellow2014generative}. In \citep{goodfellow2014distinguishability}, it is shown that the NCE objective of \cite{gutmann2010noise} can lead to maximum likelihood learning, but not the adversarial objective.

\section{Method}
\label{sec:method}
\vspace{-5pt}

The key idea of contrastive learning is very general: learn a representation that is close in some metric space for ``positive'' pairs and push apart the representation between ``negative'' pairs. 
Fig. \ref{fig:methods_diagram} gives a visual explanation for how we structure contrastive learning for the three tasks we consider: model compression, cross-modal transfer and ensemble distillation. 

\subsection{Contrastive Loss}
Given two deep neural networks, a teacher $f^T$ and a student $f^S$. Let $x$ be the network input; we denote representations at the penultimate layer (before logits) as $f^T(x)$ and $f^S(x)$ respectively. Let $x_i$ represent a training sample, and $x_j$ another randomly chosen sample. We would like to push closer the representations $f^S(x_i)$ and $f^T(x_i)$ while pushing apart $f^S(x_i)$ and $f^T(x_j)$. For ease of notation, we define random variables $S$ and $T$ for the student and teacher's representations of the data respectively:
\begin{align}
    x &\sim p_{\texttt{data}}(x) &&\triangleleft \quad \text{{\bf data}}\\
    S &= f^S(x) &&\triangleleft \quad \text{{\bf student's representation}}\\
    T &= f^T(x) &&\triangleleft \quad \text{{\bf teacher's representation}}
\end{align}
Intuitively speaking, we will consider the joint distribution $p(S,T)$ and the product of marginal distributions $p(S)p(T)$, so that, by maximizing KL divergence between these distributions, we can maximize the \emph{mutual information} between student and teacher representations. To setup an appropriate loss that can achieve this aim, let us define a distribution $q$ with latent variable $C$ which decides whether a tuple $(f^T(x_i), f^S(x_j))$ was drawn from the joint ($C=1$) or product of marginals ($C=0$):
\begin{equation}
    q(T, S | C = 1) = p(T, S), \quad q(T, S | C = 0) = p(T)p(S)
\end{equation}
Now, suppose in our data, we are given $1$ congruent pair (drawn from the joint distribution, i.e. the same input provided to $T$ and $S$) for every $N$ incongruent pairs (drawn from the product of marginals; independent randomly drawn inputs provided to $T$ and $S$). Then the priors on the latent $C$ are:
\begin{equation}
    q(C = 1) = \frac{1}{N + 1}, \quad q(C = 0) = \frac{N}{N + 1}
\end{equation}
By simple manipulation and Bayes' rule, the posterior for class $C=1$ is given by:
\begin{eqnarray}
    q(C = 1 | T, S) &=& \frac{q(T, S|C=1)q(C=1)}{q(T, S|C=0)q(C=0) + q(x, y|C=1)q(C=1)} \\
    &=& \frac{p(T, S)}{p(T, S) + N p(T)p(S)}
\end{eqnarray}
Next, we observe a connection to mutual information as follows:%
\begin{eqnarray}
    \log q(C = 1 | T, S) &=& \log \frac{p(T, S)}{p(T, S) + N p(T)p(S)} \nonumber \\
    &=& -\log (1 + N \frac{p(T)p(S)}{p(T, S)}) \leq -\log(N) + \log \frac{p(T, S)}{p(T)p(S)} \label{eqn:mutinf}
\end{eqnarray}
Then taking expectation on both sides w.r.t. $p(T, S)$ (equivalently w.r.t. $q(T, S| C = 1)$) and rearranging, gives us: 
\begin{equation}
    I(T; S) \geq \log(N) + \mathbb{E}_{q(T, S | C = 1)} \log q(C = 1 | T, S) \quad\quad \triangleleft \quad \text{{\bf MI bound}}
    \label{eqn:mi1}
\end{equation}
where $I(T; S)$ is the mutual information between the distributions of the teacher and student embeddings. Thus maximizing $\mathbb{E}_{q(T, S| C = 1)} \log q(C = 1 | T, S)$ w.r.t. the parameters of the student network $S$ increases a lower bound on mutual information. 
However, we do not know the true distribution $q(C = 1 | T, S)$; instead we estimate it by fitting a model $h: \{\mathcal{T}, \mathcal{S}\} \rightarrow [0,1]$
to samples from the data distribution $q(T, S | C=1)$ and $q(T, S | C=0)$, where $\mathcal{T}$ and $\mathcal{S}$ represent the domains of the embeddings. We maximize the log likelihood of the data under this model (a binary classification problem):
\begin{align}
    \mathcal{L}_{\textit{critic}}(h) &= \mathbb{E}_{q(T, S | C=1)} [\log h(T, S)] + N\mathbb{E}_{q(T, S | C=0)} [\log(1 - h(T, S))]\label{eqn:critic}\\
    h^* &= \argmax_h \mathcal{L}_{\textit{critic}}(h) \quad\quad\quad\quad\quad\quad\quad\quad\quad\quad\quad\quad \triangleleft \quad \text{{\bf optimal critic}}
\end{align}
We term $h$ the \emph{critic} since we will be learning representations that optimize the critic's score. Assuming sufficiently expressive $h$, $h^*(T,S) = q(C=1|T,S)$ (via Gibbs' inequality; see Sec. \ref{sec:proof} for proof), so we can rewrite Eq. \ref{eqn:mi1} in terms of $h^*$:
\begin{equation}
    I(T; S) \geq \log(N) + \mathbb{E}_{q(T, S|C=1)} [\log h^*(T, S)] 
    \label{eqn:mi2_}
\end{equation}
Therefore, we see that the optimal critic is an estimator whose expectation lower-bounds mutual information. %
We wish to learn a student that maximizes the mutual information between its representation and the teacher's, suggesting the following optimization problem:
\begin{equation}
    f^{S*} = \argmax_{f^S} \mathbb{E}_{q(T, S|C=1)} [\log h^*(T, S)]
\end{equation}
An apparent difficulty here is that the optimal critic $h^*$ depends on the current student. We can circumvent this difficulty by weakening the bound in \eqref{eqn:mi2_} to:
\begin{align}
    I(T; S) &\geq \log(N) + \mathbb{E}_{q(T, S|C=1)} [\log h^*(T, S)] + N\mathbb{E}_{q(T, S|C=0)} [\log(1-h^*(T, S))]\\
    &= \log(N) + \mathcal{L}_{\textit{critic}}(h^*) = \log(N) + \max_h \mathcal{L}_{\textit{critic}}(h)\label{eqn:opt_critic_bound}\\
    &\geq \log(N) + \mathcal{L}_{\textit{critic}}(h)\label{eqn:opt_critic_bound_weak}
\end{align}
The first line comes about by simply adding $N\mathbb{E}_{q(T, S|C=0)} [\log(1-h^*(T, S))]$ to the bound in \eqref{eqn:mi2_}. This term is strictly negative, so the inequality holds. The last line follows from the fact that $\mathcal{L}_{\textit{critic}}(h^*)$ upper-bounds $\mathcal{L}_{\textit{critic}}(h)$. Optimizing \eqref{eqn:opt_critic_bound} w.r.t. the student we have:
\begin{align}
    f^{S*} &= \argmax_{f^S}\max_h \mathcal{L}_{\textit{critic}}(h) \quad\quad\quad\quad\quad\quad\quad\quad\quad \triangleleft \quad \text{{\bf our final learning problem}}\\
    &= \argmax_{f^S}\max_h \mathbb{E}_{q(T, S|C=1)} [\log h(T, S)] + N\mathbb{E}_{q(T, S|C=0)} [\log(1-h(T, S))]
    \label{eqn:loss_NCE}
\end{align}
which demonstrates that we may jointly optimize $f^{S}$ at the same time as we learn $h$. We note that due to \eqref{eqn:opt_critic_bound_weak}, 
$f^{S*} = \argmax_{f^S}\mathcal{L}_{\textit{critic}}(h)$, 
for any $h$, \emph{also} is a representation that optimizes a lower-bound (a weaker one) on mutual information, so our formulation does not rely on $h$ being optimized perfectly.

We may choose to represent $h$ with any family of functions that satisfy $h: \{\mathcal{T}, \mathcal{S}\} \rightarrow [0,1]$. In practice, we use the following:
\begin{equation}
    h(T, S) = \frac{e^{g^T(T)^{\prime}g^S(S)/\tau}}{e^{g^T(T)^{\prime}g^S(S)/\tau} + \frac{N}{M}}
    \label{eqn:h}
\end{equation}
where $M$ is the cardinality of the dataset and $\tau$ is a temperature that adjusts the concentration level. %
In practice, since the dimensionality of $S$ and $T$ may be different, $g^S$ and $g^T$ linearly transform them into the same dimension and further normalize them by $\mathcal{L}$-2 norm before the inner product. 
The form of Eq. \eqref{eqn:loss_NCE} is inspired by NCE \citep{gutmann2010noise,wu2018unsupervised}. Our formulation is similar to the InfoNCE loss \citep{oord2018representation} in that we maximize a lower bound on the mutual information. However we use a different objective and bound, which in our experiments 
we found to be more effective than InfoNCE. 

\textbf{Implementation.} Theoretically, larger $N$ in Eq.~\ref{eqn:opt_critic_bound_weak} leads to tighter lower bound on MI. In practice, to avoid using very large batch size, we follow ~\cite{wu2018unsupervised} and implement a memory buffer that stores latent features of each data sample computed from previous batches. Therefore, during training we can efficiently retrieve a large number of negative samples from the memory buffer.

\subsection{Knowledge Distillation Objective}

The knowledge distillation loss was proposed in~\cite{hinton2015distilling}. In addition to the regular cross-entropy loss between the student output $y^S$ and one-hot label $y$, it asks the student network output to be as similar as possible to the teacher output by minimizing the cross-entropy between their output probabilities. The complete objective is:
\vspace{-2pt}
\begin{equation}
    \mathcal{L}_{KD} = (1-\alpha) H(y, y^S) + \alpha \rho^2 H(\sigma(z^T/\rho), \sigma(z^S/\rho))
    \label{eqn:kd}
\end{equation}
where $\rho$ is the temperature, $\alpha$ is a balancing weight, and $\sigma$ is softmax function. $H(\sigma(z^T/\rho), \sigma(z^S/\rho))$ is further decomposed into $KL(\sigma(z^T/\rho) | \sigma(z^S/\rho))$ and a constant entropy $H(\sigma(z^T/\rho))$.

\subsection{Cross-Modal Transfer Loss}
\vspace{-5pt}
In the cross-modal transfer task shown in Fig. ~\ref{fig:methods_diagram}(b), a teacher network is trained on a source modality $\mathcal{X}$ with large-scale labeled dataset. We then wish to transfer the knowledge to a student network, but adapt it to another dataset or modality $\mathcal{Y}$. But the features of the teacher network are still valuable to help with learning of the student on another domain. In this transfer task, we use the contrastive loss Eq.~\ref{eqn:critic} to match the features of the student and teacher. Additionally, we also consider other distillation objectives, such as KD discussed in previous section, Attention Transfer~\cite{zagoruyko2016paying} and FitNet~\cite{romero2014fitnets}. Such transfer is conducted on a paired but unlabeled dataset $D=\{(x_i, y_i)|i=1,...,L,x_i\in \mathcal{X}, y_i\in\mathcal{Y}\}$. In this scenario, there is no true label $y$ of such data for the original training task on the source modality, and therefore we ignore the $H(y,y^S)$ term in all objectives that we test. Prior cross-modal work \cite{aytar2016soundnet,hoffman2016cross,hoffman2016learning} uses either $L_2$ regression or KL-divergence.
\vspace{-5pt}

\subsection{Ensemble Distillation Loss}
\vspace{-5pt}
In the case of ensemble distillation shown in~\ref{fig:methods_diagram}(c), we have $M > 1$ teacher networks, ${f^{T_i}}$ and one student network $f^S$. We adopt the contrastive framework by defining multiple pair-wise contrastive losses %
between features of each teacher network $f^{T_i}$ and the student network $f^S$. These losses are summed together to give the final loss (to be minimized): 
\vspace{-5pt}
\begin{equation}
    \mathcal{L}_{CRD-EN} = H(y, y^S) - \beta \sum_i\mathcal{L}_{critic}(T_i, S)
    \label{eqn:en_ckd}
\end{equation}

\section{Experiments}
\label{sec:results}
\vspace{-5pt}

We evaluate our contrastive representation distillation (CRD) framework in three knowledge distillation tasks: (a) model compression of a large network to a smaller one; (b) cross-modal knowledge transfer; (c) ensemble distillation from a group of teachers to a single student network.

\textbf{Datasets} (1) \emph{CIFAR-100} \citep{krizhevsky2009learning} contains 50K training images with 0.5K images per class and 10K test images. (2) \emph{ImageNet} \citep{deng2009imagenet} provides 1.2 million images from 1K classes for training and 50K for validation. (3) \emph{STL-10} \citep{coates2011analysis} consists of a training set of 5K labeled images from 10 classes and 100K unlabeled images, and a test set of 8K images. (4) \emph{TinyImageNet} \citep{deng2009imagenet} has 200 classes, each with 500 training images and 50 validaton images. (5) \emph{NYU-Depth V2} \citep{silberman2012nyu} consists of 1449 indoor images, each labeled with dense depth image and semantic map.

\vspace{-5pt}

\subsection{Model Compression}
\label{sec:model_compression}
\vspace{-5pt}

\begin{table}[ht]

\setlength{\tabcolsep}{4.5pt}
\begin{center}
\begin{tabular}{lccccccc}
\toprule
\thead{Teacher \\ Student} & \thead{WRN-40-2 \\ WRN-16-2} & \thead{WRN-40-2 \\ WRN-40-1} & \thead{resnet56\\resnet20} & \thead{resnet110\\resnet20} & \thead{resnet110\\resnet32} & \thead{resnet32x4\\resnet8x4} & \thead{vgg13\\vgg8}\\
\midrule
Teacher & 75.61 & 75.61 & 72.34 & 74.31 & 74.31 & 79.42 & 74.64 \\
Student & 73.26 & 71.98 & 69.06 & 69.06 & 71.14 & 72.50 & 70.36 \\
\midrule

KD$^*$ & 74.92 & 73.54 & 70.66 & 70.67 & 73.08 & 73.33 & 72.98 \\
FitNet$^*$ & 73.58 ($\mathcolor{red}{\downarrow}$)& 72.24 ($\mathcolor{red}{\downarrow}$)& 69.21 ($\mathcolor{red}{\downarrow}$)& 68.99 ($\mathcolor{red}{\downarrow}$)& 71.06 ($\mathcolor{red}{\downarrow}$)& 73.50 ($\mathcolor{green}{\uparrow}$)& 71.02 ($\mathcolor{red}{\downarrow}$)\\
AT & 74.08 ($\mathcolor{red}{\downarrow}$)& 72.77 ($\mathcolor{red}{\downarrow}$)& 70.55 ($\mathcolor{red}{\downarrow}$)& 70.22 ($\mathcolor{red}{\downarrow}$)& 72.31 ($\mathcolor{red}{\downarrow}$)& 73.44 ($\mathcolor{green}{\uparrow}$)& 71.43 ($\mathcolor{red}{\downarrow}$)\\
SP & 73.83 ($\mathcolor{red}{\downarrow}$)& 72.43 ($\mathcolor{red}{\downarrow}$)& 69.67 ($\mathcolor{red}{\downarrow}$)& 70.04 ($\mathcolor{red}{\downarrow}$)& 72.69 ($\mathcolor{red}{\downarrow}$)& 72.94 ($\mathcolor{red}{\downarrow}$)& 72.68 ($\mathcolor{red}{\downarrow}$)\\
CC & 73.56 ($\mathcolor{red}{\downarrow}$)& 72.21 ($\mathcolor{red}{\downarrow}$)& 69.63 ($\mathcolor{red}{\downarrow}$)& 69.48 ($\mathcolor{red}{\downarrow}$)& 71.48 ($\mathcolor{red}{\downarrow}$)& 72.97 ($\mathcolor{red}{\downarrow}$)& 70.71 ($\mathcolor{red}{\downarrow}$)\\
VID & 74.11 ($\mathcolor{red}{\downarrow}$)& 73.30 ($\mathcolor{red}{\downarrow}$)& 70.38 ($\mathcolor{red}{\downarrow}$)& 70.16 ($\mathcolor{red}{\downarrow}$)& 72.61 ($\mathcolor{red}{\downarrow}$)& 73.09 ($\mathcolor{red}{\downarrow}$)& 71.23 ($\mathcolor{red}{\downarrow}$)\\
RKD & 73.35 ($\mathcolor{red}{\downarrow}$)& 72.22 ($\mathcolor{red}{\downarrow}$)& 69.61 ($\mathcolor{red}{\downarrow}$)& 69.25 ($\mathcolor{red}{\downarrow}$)& 71.82 ($\mathcolor{red}{\downarrow}$)& 71.90 ($\mathcolor{red}{\downarrow}$)& 71.48 ($\mathcolor{red}{\downarrow}$)\\
PKT & 74.54 ($\mathcolor{red}{\downarrow}$)& 73.45 ($\mathcolor{red}{\downarrow}$)& 70.34 ($\mathcolor{red}{\downarrow}$)& 70.25 ($\mathcolor{red}{\downarrow}$)& 72.61 ($\mathcolor{red}{\downarrow}$)& 73.64 ($\mathcolor{green}{\uparrow}$)& 72.88 ($\mathcolor{red}{\downarrow}$)\\
AB & 72.50 ($\mathcolor{red}{\downarrow}$)& 72.38 ($\mathcolor{red}{\downarrow}$)& 69.47 ($\mathcolor{red}{\downarrow}$)& 69.53 ($\mathcolor{red}{\downarrow}$)& 70.98 ($\mathcolor{red}{\downarrow}$)& 73.17 ($\mathcolor{red}{\downarrow}$)& 70.94 ($\mathcolor{red}{\downarrow}$)\\
FT$^*$ & 73.25 ($\mathcolor{red}{\downarrow}$)& 71.59 ($\mathcolor{red}{\downarrow}$)& 69.84 ($\mathcolor{red}{\downarrow}$)& 70.22 ($\mathcolor{red}{\downarrow}$)& 72.37 ($\mathcolor{red}{\downarrow}$)& 72.86 ($\mathcolor{red}{\downarrow}$)& 70.58 ($\mathcolor{red}{\downarrow}$)\\
FSP$^*$ & 72.91 ($\mathcolor{red}{\downarrow}$)& 
 n/a & 69.95 ($\mathcolor{red}{\downarrow}$)& 70.11 ($\mathcolor{red}{\downarrow}$)& 71.89 ($\mathcolor{red}{\downarrow}$)& 72.62 ($\mathcolor{red}{\downarrow}$)& 70.23 ($\mathcolor{red}{\downarrow}$)\\
NST$^*$ & 73.68 ($\mathcolor{red}{\downarrow}$)& 72.24 ($\mathcolor{red}{\downarrow}$)& 69.60 ($\mathcolor{red}{\downarrow}$)& 69.53 ($\mathcolor{red}{\downarrow}$)& 71.96 ($\mathcolor{red}{\downarrow}$)& 73.30 ($\mathcolor{red}{\downarrow}$)& 71.53 ($\mathcolor{red}{\downarrow}$)\\
CRD & \textbf{75.48} ($\mathcolor{green}{\uparrow}$)& \textbf{74.14} ($\mathcolor{green}{\uparrow}$)& \textbf{71.16} ($\mathcolor{green}{\uparrow}$)& \textbf{71.46} ($\mathcolor{green}{\uparrow}$)& \textbf{73.48} ($\mathcolor{green}{\uparrow}$)& \textbf{75.51} ($\mathcolor{green}{\uparrow}$)& \textbf{73.94} ($\mathcolor{green}{\uparrow}$)\\
\midrule
CRD+KD & 75.64 ($\mathcolor{green}{\uparrow}$)& 74.38 ($\mathcolor{green}{\uparrow}$)& 71.63 ($\mathcolor{green}{\uparrow}$)& 71.56 ($\mathcolor{green}{\uparrow}$)& 73.75 ($\mathcolor{green}{\uparrow}$)& 75.46 ($\mathcolor{green}{\uparrow}$)& 74.29 ($\mathcolor{green}{\uparrow}$)\\
\end{tabular}
\vspace{-5pt}
\caption{
\small{
Test \emph{accuracy} (\%) of student networks on CIFAR100 of a number of distillation methods (ours is CRD); see Appendix for citations of other methods. $\mathcolor{green}{\uparrow}$ denotes outperformance over KD and $\mathcolor{red}{\downarrow}$ denotes underperformance. We note that CRD is the \emph{only} method to always outperform KD (and also outperforms all other methods). We denote by * methods where we used our reimplementation based on the paper; for all other methods we used author-provided or author-verified code. Average over 5 runs.
}}
\label{tbl:cifar100_same}
\end{center}
\vspace{-10pt}
\end{table}
\begin{table}[ht]

\setlength{\tabcolsep}{4.5pt}
\begin{center}
\begin{tabular}{lcccccc}
\toprule
\thead{Teacher \\ Student} & \thead{vgg13 \\ MobileNetV2} & \thead{ResNet50 \\ MobileNetV2} & \thead{ResNet50 \\ vgg8} & \thead{resnet32x4\\ShuffleNetV1} & \thead{resnet32x4\\ShuffleNetV2} & \thead{WRN-40-2\\ShuffleNetV1}\\
\midrule
Teacher & 74.64	& 79.34 & 79.34	& 79.42	& 79.42	& 75.61 \\
Student & 64.6 & 64.6 & 70.36 & 70.5 & 71.82 & 70.5 \\
\midrule

KD$^*$ & 67.37 & 67.35 & 73.81 & 74.07 & 74.45 & 74.83 \\
FitNet$^*$ & 64.14 ($\mathcolor{red}{\downarrow}$)& 63.16 ($\mathcolor{red}{\downarrow}$)& 70.69 ($\mathcolor{red}{\downarrow}$)& 73.59 ($\mathcolor{red}{\downarrow}$)& 73.54 ($\mathcolor{red}{\downarrow}$)& 73.73 ($\mathcolor{red}{\downarrow}$)\\
AT & 59.40 ($\mathcolor{red}{\downarrow}$)& 58.58 ($\mathcolor{red}{\downarrow}$)& 71.84 ($\mathcolor{red}{\downarrow}$)& 71.73 ($\mathcolor{red}{\downarrow}$)& 72.73 ($\mathcolor{red}{\downarrow}$)& 73.32 ($\mathcolor{red}{\downarrow}$)\\
SP & 66.30 ($\mathcolor{red}{\downarrow}$)& 68.08 ($\mathcolor{green}{\uparrow}$)& 73.34 ($\mathcolor{red}{\downarrow}$)& 73.48 ($\mathcolor{red}{\downarrow}$)& 74.56 ($\mathcolor{green}{\uparrow}$)& 74.52 ($\mathcolor{red}{\downarrow}$)\\
CC & 64.86 ($\mathcolor{red}{\downarrow}$)& 65.43 ($\mathcolor{red}{\downarrow}$)& 70.25 ($\mathcolor{red}{\downarrow}$)& 71.14 ($\mathcolor{red}{\downarrow}$)& 71.29 ($\mathcolor{red}{\downarrow}$)& 71.38 ($\mathcolor{red}{\downarrow}$)\\
VID & 65.56 ($\mathcolor{red}{\downarrow}$)& 67.57 ($\mathcolor{green}{\uparrow}$)& 70.30 ($\mathcolor{red}{\downarrow}$)& 73.38 ($\mathcolor{red}{\downarrow}$)& 73.40 ($\mathcolor{red}{\downarrow}$)& 73.61 ($\mathcolor{red}{\downarrow}$)\\
RKD & 64.52 ($\mathcolor{red}{\downarrow}$)& 64.43 ($\mathcolor{red}{\downarrow}$)& 71.50 ($\mathcolor{red}{\downarrow}$)& 72.28 ($\mathcolor{red}{\downarrow}$)& 73.21 ($\mathcolor{red}{\downarrow}$)& 72.21 ($\mathcolor{red}{\downarrow}$)\\
PKT & 67.13 ($\mathcolor{red}{\downarrow}$)& 66.52 ($\mathcolor{red}{\downarrow}$)& 73.01 ($\mathcolor{red}{\downarrow}$)& 74.10 ($\mathcolor{green}{\uparrow}$)& 74.69 ($\mathcolor{green}{\uparrow}$)& 73.89 ($\mathcolor{red}{\downarrow}$)\\
AB & 66.06 ($\mathcolor{red}{\downarrow}$)& 67.20 ($\mathcolor{red}{\downarrow}$)& 70.65 ($\mathcolor{red}{\downarrow}$)& 73.55 ($\mathcolor{red}{\downarrow}$)& 74.31 ($\mathcolor{red}{\downarrow}$)& 73.34 ($\mathcolor{red}{\downarrow}$)\\
FT$^*$ & 61.78 ($\mathcolor{red}{\downarrow}$)& 60.99 ($\mathcolor{red}{\downarrow}$)& 70.29 ($\mathcolor{red}{\downarrow}$)& 71.75 ($\mathcolor{red}{\downarrow}$)& 72.50 ($\mathcolor{red}{\downarrow}$)& 72.03 ($\mathcolor{red}{\downarrow}$)\\
NST$^*$ & 58.16 ($\mathcolor{red}{\downarrow}$)& 64.96 ($\mathcolor{red}{\downarrow}$)& 71.28 ($\mathcolor{red}{\downarrow}$)& 74.12 ($\mathcolor{green}{\uparrow}$)& 74.68 ($\mathcolor{green}{\uparrow}$)& 74.89 ($\mathcolor{green}{\uparrow}$)\\
CRD & \textbf{69.73} ($\mathcolor{green}{\uparrow}$)& \textbf{69.11} ($\mathcolor{green}{\uparrow}$)& \textbf{74.30} ($\mathcolor{green}{\uparrow}$)& \textbf{75.11} ($\mathcolor{green}{\uparrow}$)& \textbf{75.65} ($\mathcolor{green}{\uparrow}$)& \textbf{76.05} ($\mathcolor{green}{\uparrow}$)\\
\midrule
CRD+KD & 69.94 ($\mathcolor{green}{\uparrow}$)& 69.54 ($\mathcolor{green}{\uparrow}$)& 74.58 ($\mathcolor{green}{\uparrow}$)& 75.12 ($\mathcolor{green}{\uparrow}$)& 76.05 ($\mathcolor{green}{\uparrow}$)& 76.27 ($\mathcolor{green}{\uparrow}$)\\
\end{tabular}
\vspace{-5pt}
\caption{
\small{
Top-1 test \emph{accuracy} (\%) of student networks on CIFAR100 of a number of distillation methods (ours is CRD) for transfer across very different teacher and student architectures. CRD outperforms KD and all other methods. Importantly, some methods that require very similar student and teacher architectures perform quite poorly. E.g. FSP \citep{yim2017gift} cannot even be applied; AT \citep{ba2014deep} and FitNet \citep{zagoruyko2016paying}  perform very poorly etc. We denote by * methods where we used our reimplementation based on the paper; for all other methods we used author-provided or author-verified code.  Average over 3 runs.
}}
\label{tbl:cifar100_diff}
\end{center}
\vspace{-10pt}
\end{table}
\textbf{Setup} We experiment on CIFAR-100 and ImageNet with student-teacher combinations of various capacity, such as ResNet \citep{he2016deep} or Wide ResNet (WRN) \citep{zagoruyko2016wide}.

\textbf{Results on CIFAR100} 
Table~\ref{tbl:cifar100_same} and Table~\ref{tbl:cifar100_diff}
compare top-1 \emph{accuracies} of different distillation objectives (for details, see Section \ref{sec:other_methods}). Table~\ref{tbl:cifar100_same} investigates students and teachers of the same architectural style, while Table~\ref{tbl:cifar100_diff} focuses on students and teachers from different architectures. We observe that our loss, which we call CRD (Contrastive Representation Distillation), consistently outperforms all other distillation objectives, including the original KD (an average relative improvement of 57\%). Surprisingly, we found that KD works pretty well and none of the other methods consistently outperforms KD on their own. Another observation is that, while switching the teacher student combinations from same to different architectural styles, methods that distill intermediate representations tend to perform worse than methods that distill from the last several layers. For example, the Attention Transfer (AT) and FitNet methods even underperform the vanilla student. In contrast, PKT, SP and CRD that operate on last several layers performs well. This might because, architectures of different styles have their own solution paths mapping from the input to the output, and enforcing the mimic of intermediate representations thus might conflict with such inductive bias.

\begin{figure}[t]
\centering
\vspace{-5pt}
\includegraphics[width=\linewidth]{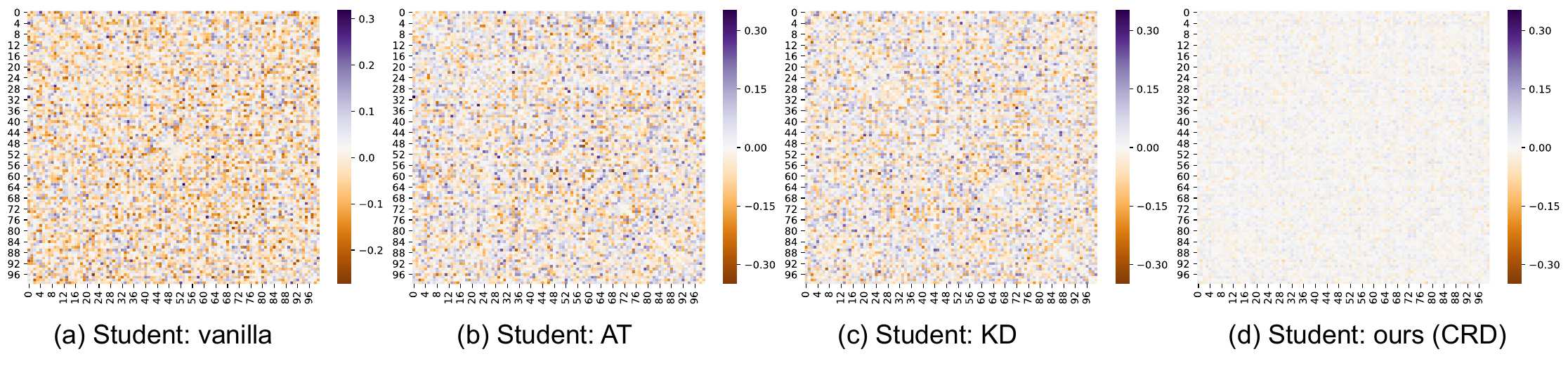}
\vspace{-20pt}
\caption{\small{
The correlations between class logits of a teacher network are ignored by regular cross-entropy. Distillation frameworks use ``soft targets" \citep{hinton2015distilling} which effectively capture such correlations and transfer them to the student network, leading to the success of distillation. We visualize here the \emph{difference} of correlation matrices of student and teacher logits, for different student networks on a CIFAR-100 knowledge distillation task: (a) Student trained without distillation, showing that the teacher and student cross-correlations are very different; (b) Student distilled by attention transfer~\citep{zagoruyko2016paying}; showing reduced difference (see axis); (c) Student distilled by KL divergence \citep{hinton2015distilling}, also showing reduced difference; (d) Student distilled by our contrastive objective, showing significant matching between student's and teacher's correlations. In this visualization, we use WRN-40-2 as teacher and WRN-40-1 as student.
}}
\label{fig:correlation}%
\end{figure}

\textbf{Capturing inter-class correlations.} In Fig.~\ref{fig:correlation}, we compute the difference between the correlation matrices of the teacher's and student's logits; for three different students: vanilla student without distillation, trained by AT, KD or CRD (our method). It is clear that the CRD objective captures the most correlation structure in the logit as shown by the smaller differences between teacher and student. This is reflected in reduced error rates.

\begin{table}[t]
\setlength{\tabcolsep}{4.5pt}
\begin{center}
\begin{tabular}{l|cc|cccccc|c}
\toprule
 & Teacher & Student & AT & KD & SP & CC & Online KD * & CRD & CRD+KD \\
\midrule
Top-1 & 26.69 & 30.25 & 29.30 & 29.34 & 29.38 & 30.04 & 29.45 & 28.83 & \textbf{28.62} \\
Top-5 & 8.58 & 10.93 & 10.00 & 10.12 & 10.20 & 10.83 & 10.41 & 9.87 & \textbf{9.51}\\
\bottomrule
\end{tabular}
\caption{
\small{Top-1 and Top-5 error rates (\%) of student network ResNet-18 on ImageNet validation set. We use ResNet-34 released by PyTorch team as our teacher network, and follow the standard training practice of ImageNet on PyTorch except that we train for 10 more epochs. We compare our CRD with KD~\citep{hinton2015distilling}, AT~\citep{zagoruyko2016paying} and Online-KD~\citep{lan2018knowledge}. ``*'' reported by the original paper~\cite{lan2018knowledge} using an ensemble of online ResNets as teacher, no pretrained ResNet-34 was used.}
}
\label{tbl:imagenet}
\end{center}
\vspace{-5pt}
\end{table}

\textbf{Results on ImageNet} For a fair comparison with \cite{zagoruyko2016paying} and \cite{lan2018knowledge}, we adopt the models from these papers, ResNet-34 as the teacher and ResNet-18 as the student. As shown in Table~\ref{tbl:imagenet}, the gap of top-1 accuracy between the teacher and student is $3.56\%$. The AT method reduces this gap by $0.95\%$, while ours narrow it by $1.42\%$, a $50\%$ relative improvement. 
Results on ImageNet validates the scalability of our CRD.

\begin{table}[t]
\setlength{\tabcolsep}{4.5pt}
\begin{center}
\begin{tabular}{l|c|ccccc|c}
\toprule
 & Student & KD & AT & FitNet & CRD & CRD+KD & Teacher\\
\midrule
CIFAR100$\rightarrow$STL-10        & 69.7 & 70.9 & 70.7 & 70.3 & 71.6 & \textbf{72.2} & 68.6 \\
CIFAR100$\rightarrow$TinyImageNet & 33.7 & 33.9 & 34.2 & 33.5 & \textbf{35.6} & 35.5 & 31.5 \\
\bottomrule
\end{tabular}
\caption{
\small{We transfer the representation learned from CIFAR100 to STL-10 and TinyImageNet datasets by freezing the network and training a linear classifier on top of the last feature layer to perform 10-way (STL-10) or 200-way (TinyImageNet) classification. For this experiment, we use the combination of teacher network WRN-40-2 and student network WRN-16-2. Classification accuracies (\%) are reported.
}
}
\label{tbl:transfer}
\end{center}
\vspace{-5pt}
\end{table}

\textbf{Transferability of representations} We are interested in \emph{representations}, and a primary goal of representation learning is to acquire \emph{general} knowledge, that is, knowledge that transfers to tasks or datasets that were unseen during training. Therefore, we test if the representations we distill transfer well. A WRN-16-2 student either distills from a WRN-40-2 teacher, or is trained from scratch on CIFAR100. The student serves as a frozen representation extractor (the layer prior to the logit) for images from STL-10 or TinyImageNet (all images downsampled to 32x32). We then train a \emph{linear} classifier to perform 10-way (for STL-10) or 200-way (for TinyImageNet) classification to quantify the transferability of the representations. We compare CRD with multiple baselines such as KD and AT in Table~\ref{tbl:transfer}. In this setting, all distillation methods except FitNet improve transferability of the learned representations on both STL-10 and TinyImageNet. While the teacher performs the best on the original CIFAR100 dataset, its representations transfer the worst to the other two datasets. This is perhaps the teacher's representations are biased towards the original task. Surprisingly, the student with CRD+KD distillation not only matches its teacher on CIFAR100 (see  Table~\ref{tbl:transfer}), but also transfers much better than the teacher, e.g., 3.6$\%$ improvement (STL-10) and 4.1$\%$ on TinyImageNet. %

\subsection{Cross-modal transfer}
\vspace{-5pt}
We consider a practical setting where modality $\mathcal{X}$ has large amount of labeled data while modality $\mathcal{Y}$ does not. Transfering knowledge from $\mathcal{X}$ to $\mathcal{Y}$ is a common challenge. For example, while large-scale RGB datasets are easily accessible, other modalities such as depth images are much harder to label at scale but have wide applications. We demonstrate the potential of CRD for cross-modal transfer in two scenarios: (a) transfer from luminance to chrominance; (b) transfer from RGB to depth images.

\begin{figure}[t]
\centering
\vspace{-0pt}
\includegraphics[width=\linewidth]{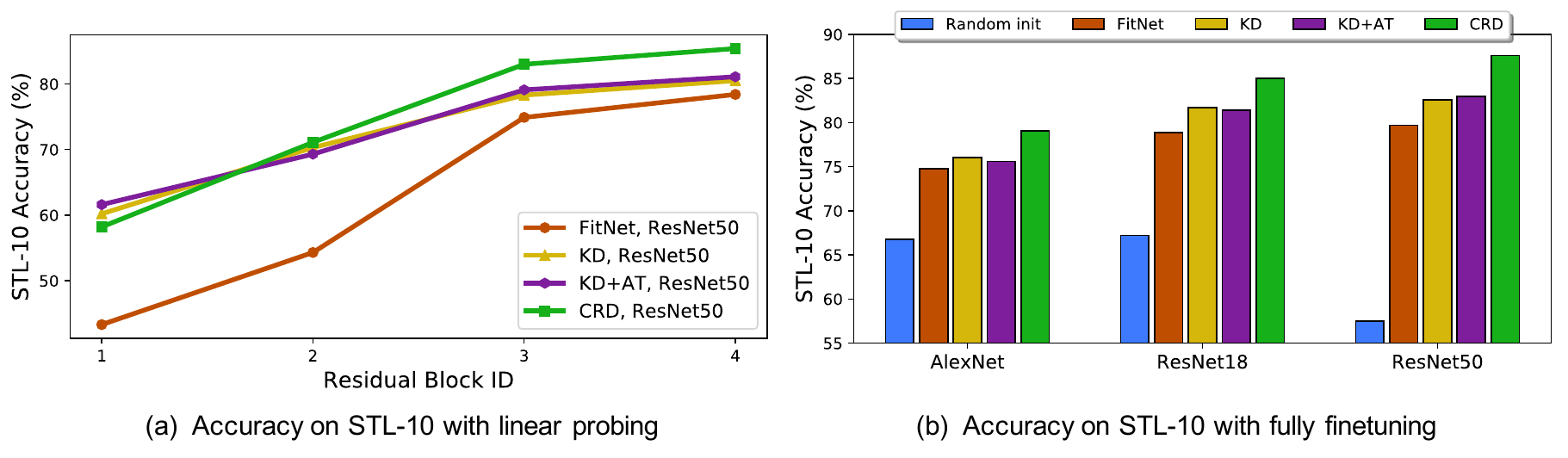}
\vspace{-20pt}
\caption{
\small{Top-1 classification accuracy on STL-10 using \emph{chrominance} image ($ab$ channel in $Lab$ color space). We initialize the \emph{chrominance} network randomly or by distilling from a \emph{luminance} network, trained with large-scale labeled images. We evaluate distillation performance by (a) linear probing and (b) fully finetuning.}
}
\label{fig:STL}\vspace{-5pt}
\end{figure}

\textbf{Transferring from Luminance to Chrominace.} We work on $Lab$ color space, where $L$ represents Luminance and $ab$ Chrominance. We first train an $L$ network on TinyImageNet with supervision. Then we transfer knowledge from this $L$ network to $ab$ network on the unlabeled set of STL-10 with different objectives, including FitNet, KD, KD+AT and CRD. For convenience, we use the same architecture for student and teacher (they can also be different). Finally, we evaluate the knowlege of $ab$ network by two means: (1) \emph{linear probing}: we freeze the $ab$ network and train a linear classifier on top of features from different layers to perform 10-way classification on STL-10 $ab$ images. This is a common practice \citep{alain2016understanding,zhang2017split} to evaluate the quality of network representations; (b) \emph{fully finetuning}: we fully finetune the $ab$ network to obtain the best accuracy. We also use as a baseline the $ab$ network that is randomly initialized rather than distilled. Architectures investigated include AlexNet, ResNet-18 and ResNet-50. The results shown in Figure~\ref{fig:STL} show that CRD is more efficient for transferring inter-modal knowledge than other methods. Besides, we also note KD+AT does not improve upon KD, possibly because attention of luminance and chrominance are different and harder to transfer.

\begin{table}[t]
\begin{center}
\begin{tabular}{l|ccccc}
\toprule
 Metric (\%) & Random Init. & KD & KD+AT & FitNet & CRD \\
\midrule
Pix. Acc. &  56.4 & 58.9 & 60.1 & 60.8 & \textbf{61.6} \\
mIoU &  35.8 & 38.0 & 39.5 & 40.7 & \textbf{41.8} \\
\bottomrule
\end{tabular}
\caption{
\small{Performance on the task of using depth to predict semantic segmentation labels. We initialize the depth network either randomly or by distilling from a ImageNet pre-pretrained ResNet-18 teacher.}
}\label{tbl:depth}
\end{center}
\vspace{-5pt}
\end{table}

\textbf{Transferring from RGB to Depth.} We transfer the knowledge of a ResNet-18 teacher pretrained on ImageNet to a 5-layer student CNN operating on depth images. We follow a similar transferring procedure on NYU-Depth training set, except that we use a trick of contrasting between local and global features, proposed by~\cite{hjelm2018learning}, to overcome the problem of insufficient data samples in the depth domain. Then the student network is further trained to predict semantic segmentation maps from depth images. We note that both knowledge transfer and downstream training are conducted on the same set of images, i.e., the training set. %
Table~\ref{tbl:depth} reports the average pixel prediction accuracy and mean Intersection-over-Union across all classes. All distillation methods can transfer knowledge from the RGB ResNet to the depth CNN. FitNet surpasses KD and KD+AT. CRD significantly outperforms all other methods.

\vspace{-5pt}
\subsection{Distillation from an ensemble}
\vspace{-5pt}

\begin{figure}[t]
\centering
\vspace{0pt}
\includegraphics[width=\linewidth]{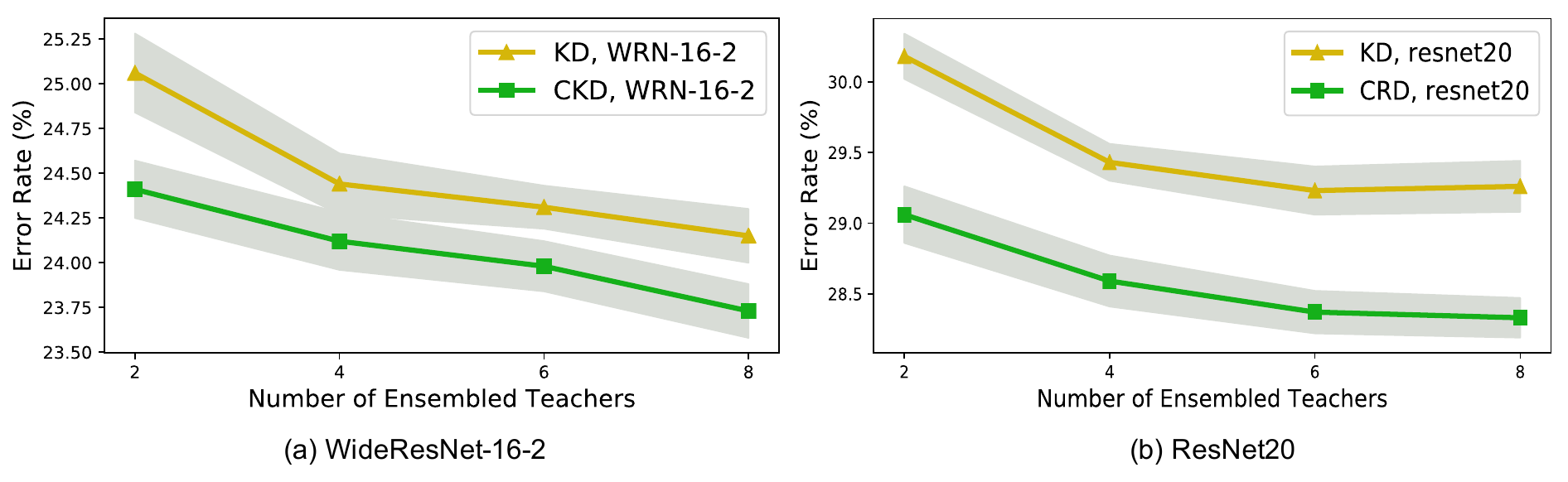}
\vspace{-10pt}
\caption{
\small{
Distillation from an ensemble of teachers. We vary the number of ensembled teachers and compare KD with our CRD by using (a) WRN-16-2 and (b) ResNet20. Our CRD consistently achieves lower error rate. 
}
}
\label{fig:ensemble}\vspace{-5pt}
\end{figure}
Better classification performance is often achieved by ensembles of deep networks, but these are usually too expensive for inference time, and distillation into a single network is a desirable task. We investigate the KL-divergence based KD and our CRD for this task, using the loss of Sec. \ref{eqn:en_ckd}. The network structures of each teacher and student are identical here, but an ensemble of multiple teachers can still provide rich knowledge for the student.  
To compare between KD and CRD on CIFAR100 dataset, we use WRN-16-2 and ResNet-20, whose single model error rates are $26.7\%$ and $30.9\%$ respectively.
The results of distillation are presented in Figure~\ref{fig:ensemble}, where we vary the number of ensembled teachers. CRD with 8 teachers decreases the error rate of WRN-16-2 to $23.7\%$ and ResNet20 to $28.3\%$. In addition, CRD works consistently better than KD in all settings we test. These observations suggest that CRD is capable of distilling an ensemble of models into a single one which performs significantly better than a model of the same size that is trained from scratch.

\vspace{-5pt}
\subsection{Ablative Study}\label{sec:ablation}

\begin{table}[t]
\begin{center}
\begin{tabular}{llccccc}
\toprule
sampling & objective & \thead{WRN-40-2 \\ WRN-16-2} & \thead{resnet110\\resnet20} & \thead{resnet110\\resnet32} & \thead{resnet32x4\\resnet8x4} & \thead{vgg13\\vgg8}\\
\midrule
\multirow{2}{*}{$i\neq j$}& 
InfoNCE & 74.78 & 70.56 & 72.67 & 74.69 & 73.24 \\
& Ours & 74.48 & 70.64 & 72.64 & 74.67 & 73.39 \\
\midrule
\multirow{2}{*}{$y_i\neq y_j$}& 
InfoNCE & 75.15 & 71.39 & \textbf{73.53} & 75.22 & 73.74 \\
& Ours & \textbf{75.48} & \textbf{71.46} & 73.48 & \textbf{75.51} & \textbf{73.94} \\
\midrule
\end{tabular}
\vspace{-5pt}
\caption{
\small{
Ablative study of different contrastive objectives and negative sampling policies on CIFAR100. For contrastive objectives, we compare our objective with InfoNCE~\citep{oord2018representation}; For negative sampling policy, when given an anchor image $x_i$ from the dataset, we consider either randomly sample negative $x_j$ such that (a) $i\neq j$, or (b) $y_i \neq y_j$ where $y$ represents the class label. Average over 5 runs.
}}
\label{tbl:cifar100_ablation}
\end{center}
\vspace{-10pt}
\end{table}
\textbf{InfoNCE v.s. Ours.} InfoNCE~\citep{oord2018representation} is an alternative contrastive objective which select a single positive out from a set of distractors via a softmax function. We compare InfoNCE with our contrastive objective Eq~\ref{eqn:loss_NCE}, when using the same number of negatives. The last two rows in Table~\ref{tbl:cifar100_ablation} show that our objective outperforms InfoNCE in 4 out of 5 teacher-student combinations.

\textbf{Negative Sampling.} In this paper, we consider two negative sampling plocies when giving an anchor $x_i$: (1) ${x_j}_{j\neq i}$ for the unsupervised case when we have no labels, or (2) ${x_j}_{y_j \neq y_i}$ for supervised case, where $y_i$ represents the label associated with sample $x_i$. One might imagine that the first sampling strategy will increase the intra-class variance as we may push apart positives and negatives from the same underlying class, while the second would not. We quantitatively measure and report the gap between these two strategies in Table~\ref{tbl:cifar100_ablation}, which shows that the classification accuracy by the second sampling strategy is $0.81\%$ higher for our objective and $0.62\%$ higher for InfoNCE than the first one.
\subsection{Hyper-parameters and computation overhead}\label{sec:parameters}

\vspace{-5pt}

\begin{figure}[t]
\centering
\includegraphics[width=\linewidth]{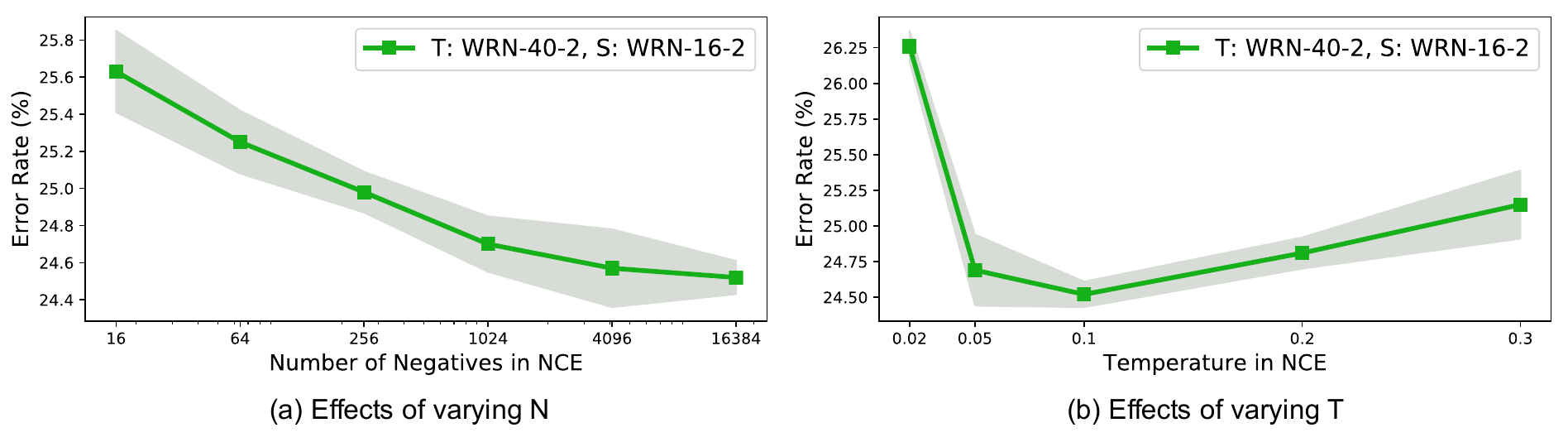}
\vspace{-15pt}
\caption{\small{
Effects of varying the number of negatives, shown in (a), or the temperature, shown in (b).
}}
\label{fig:parameter}\vspace{-10pt}
\end{figure}

There are two main hyper-parameters in our contrastive objectives: (1) the number of negative samples $N$ in Eq. \ref{eqn:loss_NCE}, and (2) temperature $\tau$ which modulates the softmax probability. We adopt WRN-40-2 as teacher and WRN-16-2 as student for parameter analysis. Experiments are conducted on CIFAR100 and the results are shown in Figure~\ref{fig:parameter}. 

\textbf{Number of negatives $N$} We have validated different $N$: $16, 64, 256, 1024, 4096, 16384$. As shown in Figure~\ref{fig:parameter}(a), increasing $N$ leads to improved performance. However, the difference of error rate between $N=4096$ and $N=16384$ is less than $0.1\%$. Therefore, we use $N=16384$ for reporting the accuracy while in practice $N=4096$ should suffice.

\textbf{Temperature $\tau$} We varied $\tau$ between $0.02$ and $0.3$. As Figure~\ref{fig:parameter}(b) illustrates, both extremely high or low temperature lead to a sub-optimal solution. In general, temperatures between $0.05$ and $0.2$ work well on CIFAR100. All experiments but those on ImageNet use a temperature of $0.1$. For ImageNet, we use $\tau=0.07$.
The optimal temperature may vary across different datasets and require further tuning.

\textbf{Computational Cost} We use ResNet-18 on ImageNet for illustration. CRD uses extra $260$ MFLOPs, which is about $12\%$ of the original $2$ GFLOPs. In practice, we did not notice significant difference of training time on ImageNet (e.g., 1.75 epochs/hour v.s. 1.67 epochs/hour on two Titan-V GPUs). The memory bank for storing all $128$-d features of ImageNet only costs around $600$MB memory, and therefore we store it on GPU memory. 

\vspace{-10pt}

\section{Conclusion}
\vspace{-5pt}
We have developed a novel technique for neural network distillation, using the concept of contrastive objectives, which are usually used for representation learning. We experimented with our objective on a number of applications such as model compression, cross-modal transfer and ensemble distillation, %
outperforming other distillation objectives by significant margins in all these tasks. Our contrastive objective is the only distillation objective that consistently outperforms knowledge distillation across a wide variety of knowledge transfer tasks. Prior objectives only surpass KD \emph{when combined} with KD. Contrastive learning is a simple and effective objective with practical benefits.
\myparagraph{Acknowledgments.} We thank Baoyun Peng for providing the code of CC~\citep{peng2019correlation} and Frederick Tung for verifying our reimplementation of SP~\citep{tung2019similarity}. This research was supported in part by Google Cloud and iFlytek.

\bibliographystyle{iclr2020_conference}
\bibliography{main}

\section{Appendix}
\label{sec:appendix}
\vspace{-5pt}

\subsection{Other Methods}
\label{sec:other_methods}
\vspace{-5pt}
We compare to the following other state-of-the-art methods from the literature:
\begin{enumerate}
    \item Knowledge Distillation (KD) \citep{hinton2015distilling}
    \item Fitnets: Hints for thin deep nets \citep{romero2014fitnets}
    \item Attention Transfer (AT) \citep{zagoruyko2016paying}
    \item Similarity-Preserving Knowledge Distillation (SP) \citep{tung2019similarity}; 
    \item Correlation Congruence (CC) \citep{peng2019correlation}
    \item Variational information distillation for knowledge transfer (VID) \citep{ahn2019variational}
    \item Relational Knowledge Distillation (RKD) \citep{park2019relational}
    \item Learning deep representations with probabilistic knowledge transfer (PKT) \citep{passalis2018learning}
    \item Knowledge transfer via distillation of activation boundaries formed by hidden neurons (AB) \citep{heo2019knowledge}
    \item Paraphrasing complex network: Network compression via factor transfer (FT) \citep{kim2018paraphrasing}
    \item A gift from knowledge distillation: Fast optimization, network minimization and transfer learning (FSP) \citep{yim2017gift}
    \item Like what you like: Knowledge distill via neuron selectivity transfer (NST) \citep{huang2017like}
\end{enumerate}

\vspace{-5pt}
\subsection{Contrastive loss -- details}

\subsubsection{Proof that $h^*(T,S) = q(C=1|T,S)$}
\label{sec:proof}

We wish to model some true distribution $q(C|T=t,S=s)$. $C$ is a binary variable, so we can model $q(C|T=t,S=s)$ as a Bernoulli distribution with a single parameter $h(S=s,T=t) \in [0,1]$, defining for convenience $h'(C=1,S=s,T=t) = h(S=s,T=t)$ and $h'(C=0,S=s,T=t) = 1-h(S=s,T=t)$. The log likelihood function is:
\begin{align}
    \mathbb{E}_{c \sim q(C|S=s,T=t)}[\log h'(C=c,S=s,T=t)]\label{eqn:bernoulli_log_like}
\end{align}
By Gibbs' inequality, the max likelihood fit is $h'(C=c,S=s,T=t) = q(C=c|S=s,T=t)$, which also implies that $h(S=s,T=t) = q(C=1|S=s,T=t)$.

We now demonstrate that our objective in Eq. \eqref{eqn:critic} is proportional to a summation over terms Eq. \eqref{eqn:bernoulli_log_like} for all $s \in \mathcal{S}$ and $t \in \mathcal{T}$.
\begin{align}
    &\mathbb{E}_{s,t \sim q(S,T)}[\mathbb{E}_{c \sim q(C|S=s,T=t)} [\log h'(C=c, S=s, T=t)]]\label{eqn:log_loss_start}\\
    &=\mathbb{E}_{c,s,t \sim q(C,S,T)}[\log h'(C=c, S=s, T=t)]]\\
    &=\mathbb{E}_{s,t \sim q(S,T|C=1)q(C=1)}[\log h(S=s,T=t)] + \nonumber \\ &\mathbb{E}_{s,t \sim q(S,T|C=0)q(C=0)}[\log (1 - h(S=s,T=t))]\\
    &=\frac{1}{N+1}\mathbb{E}_{s,t \sim q(S,T|C=1)}[\log h(S=s,T=t)] + \nonumber\\ &\frac{N}{N+1}\mathbb{E}_{s,t \sim q(S,T|C=0)}[\log (1 - h(S=s,T=t))]\label{eqn:log_loss_end}
\end{align}
Notice that \eqref{eqn:log_loss_end} is proportional to Eq. \eqref{eqn:critic} from the main paper. For sufficiently expressive $h$, then, each term inside the expectation in Eq. \eqref{eqn:log_loss_start} can be maximized, resulting in $h^*(T=t,S=s) = q(C=1|T=t,S=t)$ for all $s$ and $t$. $\qed$

\subsection{Network Architectures}

Wide Residual Network (WRN)~\citep{zagoruyko2016wide}. WRN-d-w represnets wide resnet with depth $d$ and width factor $w$.

resnet~\citep{he2016deep}. We use resnet-d to represent \textbf{cifar}-style resnet with 3 groups of basic blocks, each with $16$, $32$, and $64$ channels respectively. In our experiments, resnet8 x4 and resnet32 x4 indicate a 4 times wider network (namely, with $64$, $128$, and $256$ channels for each of the block)

ResNet~\citep{he2016deep}. ResNet-d represents \textbf{ImageNet}-style ResNet with Bottleneck blocks and more channels.

MobileNetV2~\cite{sandler2018mobilenetv2}. In our experiments, we use a width multiplier of $0.5$.

vgg~\citep{simonyan2014very}. the vgg net used in our experiments are adapted from its original ImageNet counterpart.

ShuffleNetV1~\citep{zhang2018shufflenet}, ShuffleNetV2~\citep{tan2019mnasnet}. ShuffleNets are proposed for efficient training and we adapt them to input of size 32x32.

\subsection{Implementation Details}

All methods evaluated in our experiments use SGD. 

For CIFAR-100, we initialize the learning rate as 0.05, and decay it by 0.1 every 30 epochs after the first 150 epochs until the last 240 epoch. For MobileNetV2, ShuffleNetV1 and ShuffleNetV2, we use a learning rate of 0.01 as this learning rate is optimal for these models in a grid search, while 0.05 is optimal for other models.

For ImageNet, we follow the standard PyTorch practice but train for 10 more epochs. Batch size is 64 for CIFAR-100 or 256 for ImageNet.

The student is trained by a combination of cross-entropy classification objective and a knowledge distillation objective, shown as follows:
\begin{equation}
    \mathcal{L} = \mathcal{L}_{cross-entropy} + \beta \mathcal{L}_{distill}
\end{equation}

For the weight balance factor $\beta$, we directly use the optimal value from the original paper if it is specified, or do a grid search with teacher WRN-40-2 and student WRN-16-2. This results in the following list of $\beta$ used for different objectives:
\begin{enumerate}
    \item Fitnets~\citep{romero2014fitnets}: $\beta = 100$
    \item AT~\citep{zagoruyko2016paying}: $\beta = 1000$
    \item SP~\citep{tung2019similarity}: $\beta = 3000$ 
    \item CC~\citep{peng2019correlation}: $\beta = 0.02$
    \item VID~\citep{ahn2019variational}: $\beta = 1$
    \item RKD~\citep{park2019relational}: $\beta_{1} = 25$ for distance and $\beta_{2} = 50$ for angle. For this loss, we combine both term following the original paper.
    \item PKT~\citep{passalis2018learning}: $\beta = 30000$
    \item AB~\citep{heo2019knowledge}: $\beta = 0$, distillation happens in a separate pre-training stage where only distillation objective applies.
    \item FT~\citep{kim2018paraphrasing}: $\beta = 500$
    \item FSP~\citep{yim2017gift}: $\beta = 0$, distillation happens in a separate pre-training stage where only distillation objective applies.
    \item NST~\citep{huang2017like}: $\beta = 50$
    \item CRD: $\beta = 0.8$, in general $\beta \in [0.5, 1.5]$ works reasonably well.
\end{enumerate}

For KD\citep{hinton2015distilling}, we follow Eq.~\ref{eqn:kd} and set $\alpha=0.9$ and $T=4$.

\subsection{Visualization of the correlation discrepancy}
\begin{figure}[t]
\centering
\vspace{-20pt}
\includegraphics[width=\linewidth]{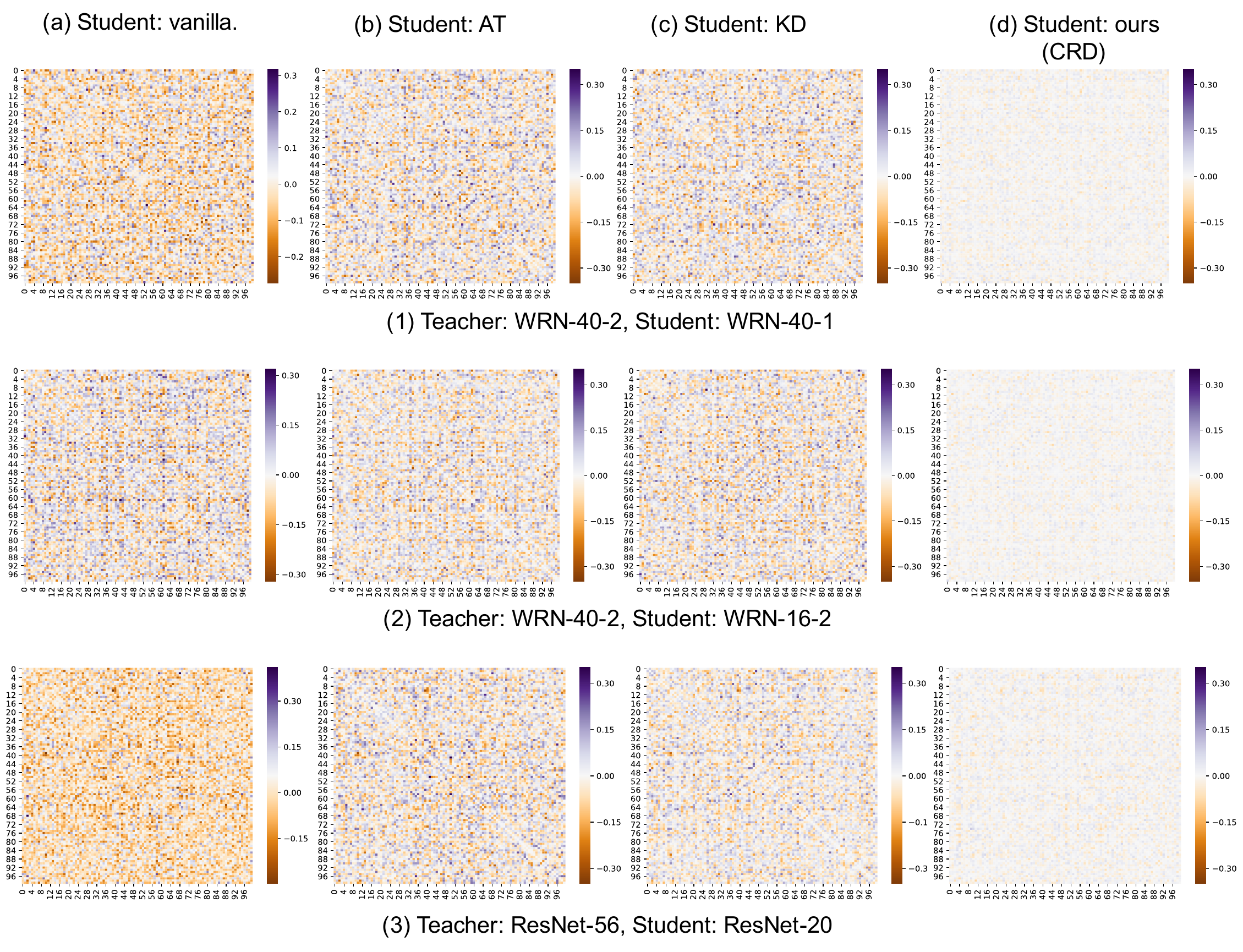}
\vspace{-10pt}
\caption{\small{The correlations between class logits output by the teacher network show the  ``dark knowledge" \cite{hinton2015distilling} that must be transferred to a student networks. A student network that captures these correlations tends to perform better at the task. We visualize here the difference between correlation matrices of the student and teacher at the logits, for different student networks on a Cifar100 knowledge distillation task: (a) A student trained without distillation; (b) A student distilled by attention transfer~\cite{zagoruyko2016paying} (c) A student distilled by KL divergence \cite{hinton2015distilling}; (d) A student distilled by our contrastive objective. Our objective greatly improves the structured knowledge (correlations) in the output units. 
}}
\label{fig:correlation_more}
\end{figure}

\vspace{-5pt}
We visualize the correlation discrepancy for different distillation objectives across various combinations of student and teacher networks. As shown in Fig.~\ref{fig:correlation_more}, our contrastive distillation objective siginificantly outperforms other objectives, in terms of minimizing the correlation discrepancy between student and teacher networks. The normalized correlation coefficients are computed at the logit layer.

\subsection{Combining different distillation objectives}

Table~\ref{tbl:cifar100_combine} shows the effect of combining various objectives with KD. Most methods is able to slightly outperform KD after this combination. We also the check the compatibility of our distillation objective with KD and PKT. The combination of CRD and KD/PKT further improves over single CRD objective. 

\begin{table}[ht]

\setlength{\tabcolsep}{4.5pt}
\begin{center}
\begin{tabular}{lccccccc}
\toprule
\thead{Teacher \\ Student} & 
\thead{WRN-40-2 \\ WRN-16-2} & 
\thead{resnet110\\resnet20} & 
\thead{resnet32x4\\resnet8x4} & 
\thead{vgg13\\vgg8} &
\thead{vgg13 \\ MobileNetV2} &
\thead{ResNet50 \\ vgg8} &
\thead{resnet32x4\\ShuffleNetV2}\\
\midrule
Teacher & 75.61 & 74.31 & 79.42 & 74.64 & 74.64 & 79.34 & 79.42 \\
Student & 73.26 & 69.06 & 72.50 & 70.36 & 64.60& 70.36 & 71.82 \\
\midrule

KD & 74.92 & 70.67 & 73.33 & 72.98 & 67.37 & 73.81 & 74.45 \\
FitNet+KD & 75.12 ($\mathcolor{green}{\uparrow}$)& 70.67 ($\mathcolor{green}{\uparrow}$)& 74.66 ($\mathcolor{green}{\uparrow}$)& 73.22 ($\mathcolor{green}{\uparrow}$)& 66.90 ($\mathcolor{red}{\downarrow}$)& 73.24 ($\mathcolor{red}{\downarrow}$)& 75.15 ($\mathcolor{green}{\uparrow}$)\\
AT+KD & 75.32 ($\mathcolor{green}{\uparrow}$)& 70.97 ($\mathcolor{green}{\uparrow}$)& 74.53 ($\mathcolor{green}{\uparrow}$)& 73.48 ($\mathcolor{green}{\uparrow}$)& 65.13 ($\mathcolor{red}{\downarrow}$)& 74.01 ($\mathcolor{green}{\uparrow}$)& 75.39 ($\mathcolor{green}{\uparrow}$)\\
SP+KD & 74.98 ($\mathcolor{green}{\uparrow}$)& 71.02 ($\mathcolor{green}{\uparrow}$)& 74.02 ($\mathcolor{green}{\uparrow}$)& 73.49 ($\mathcolor{green}{\uparrow}$)& 68.41 ($\mathcolor{green}{\uparrow}$)& 73.52 ($\mathcolor{red}{\downarrow}$)& 74.88 ($\mathcolor{green}{\uparrow}$)\\
CC+KD & 75.09 ($\mathcolor{green}{\uparrow}$)& 70.88 ($\mathcolor{green}{\uparrow}$)& 74.21 ($\mathcolor{green}{\uparrow}$)& 73.04 ($\mathcolor{green}{\uparrow}$)& 68.02 ($\mathcolor{green}{\uparrow}$)& 73.48 ($\mathcolor{red}{\downarrow}$)& 74.71 ($\mathcolor{green}{\uparrow}$)\\
VID+KD & 75.14 ($\mathcolor{green}{\uparrow}$)& 71.10 ($\mathcolor{green}{\uparrow}$)& 74.56 ($\mathcolor{green}{\uparrow}$)& 73.19 ($\mathcolor{green}{\uparrow}$)& 68.27 ($\mathcolor{green}{\uparrow}$)& 73.46 ($\mathcolor{red}{\downarrow}$)& 74.85 ($\mathcolor{green}{\uparrow}$)\\
RKD+KD & 74.89 ($\mathcolor{red}{\downarrow}$)& 70.77 ($\mathcolor{green}{\uparrow}$)& 73.79 ($\mathcolor{green}{\uparrow}$)& 72.97 ($\mathcolor{red}{\downarrow}$)& 67.87 ($\mathcolor{green}{\uparrow}$)& 73.51 ($\mathcolor{red}{\downarrow}$)& 74.55 ($\mathcolor{green}{\uparrow}$)\\
PKT+KD & 75.33 ($\mathcolor{green}{\uparrow}$)& 70.72 ($\mathcolor{green}{\uparrow}$)& 74.23 ($\mathcolor{green}{\uparrow}$)& 73.25 ($\mathcolor{green}{\uparrow}$)& 68.13 ($\mathcolor{green}{\uparrow}$)& 73.61 ($\mathcolor{red}{\downarrow}$)& 74.66 ($\mathcolor{green}{\uparrow}$)\\
AB+KD & 70.27 ($\mathcolor{red}{\downarrow}$)& 70.97 ($\mathcolor{green}{\uparrow}$)& 74.40 ($\mathcolor{green}{\uparrow}$)& 73.35 ($\mathcolor{green}{\uparrow}$)& 68.23 ($\mathcolor{green}{\uparrow}$)& 73.65 ($\mathcolor{red}{\downarrow}$)& 74.99 ($\mathcolor{green}{\uparrow}$)\\
FT+KD & 75.15 ($\mathcolor{green}{\uparrow}$)& 70.88 ($\mathcolor{green}{\uparrow}$)& 74.62 ($\mathcolor{green}{\uparrow}$)& 73.44 ($\mathcolor{green}{\uparrow}$)& 66.99 ($\mathcolor{red}{\downarrow}$)& 72.98 ($\mathcolor{red}{\downarrow}$)& 75.06 ($\mathcolor{green}{\uparrow}$)\\
NST+KD & 74.67 ($\mathcolor{red}{\downarrow}$)& 71.01 ($\mathcolor{green}{\uparrow}$)& 74.28 ($\mathcolor{green}{\uparrow}$)& 73.33 ($\mathcolor{green}{\uparrow}$)& 63.77 ($\mathcolor{red}{\downarrow}$)& 71.74 ($\mathcolor{red}{\downarrow}$)& 75.24 ($\mathcolor{green}{\uparrow}$)\\
\midrule
CRD & 75.48 ($\mathcolor{green}{\uparrow}$)& 71.46 ($\mathcolor{green}{\uparrow}$)& 75.51 ($\mathcolor{green}{\uparrow}$)& 73.94 ($\mathcolor{green}{\uparrow}$)& 69.73 ($\mathcolor{green}{\uparrow}$)& 74.30 ($\mathcolor{green}{\uparrow}$)& 75.65 ($\mathcolor{green}{\uparrow}$)\\
CRD+KD & 75.64 ($\mathcolor{green}{\uparrow}$)& 71.56 ($\mathcolor{green}{\uparrow}$)& 75.46 ($\mathcolor{green}{\uparrow}$)& 74.29 ($\mathcolor{green}{\uparrow}$)& \textbf{69.94} ($\mathcolor{green}{\uparrow}$)& 74.58 ($\mathcolor{green}{\uparrow}$)& \textbf{76.05} ($\mathcolor{green}{\uparrow}$)\\
CRD+PKT & \textbf{75.91} ($\mathcolor{green}{\uparrow}$)& \textbf{71.65} ($\mathcolor{green}{\uparrow}$)& \textbf{75.90} ($\mathcolor{green}{\uparrow}$)& \textbf{74.57} ($\mathcolor{green}{\uparrow}$)& 69.59 ($\mathcolor{green}{\uparrow}$)& \textbf{74.68} ($\mathcolor{green}{\uparrow}$)& 75.94 ($\mathcolor{green}{\uparrow}$)\\

\end{tabular}
\vspace{-5pt}
\caption{
\small{
Test \emph{accuracy} (\%) of student networks on CIFAR100 of combining distillation methods with KD; we check the compatibility of our objective with KD as well as PKT. $\mathcolor{green}{\uparrow}$ denotes outperformance over KD and $\mathcolor{red}{\downarrow}$ denotes underperformance.
}}
\label{tbl:cifar100_combine}
\end{center}
\vspace{-10pt}
\end{table}

\subsection{Additional results on Transferability of representations}

In addtion to Table~\ref{tbl:transfer} shown in the main text, we also evaluate the transferabilit of the representations of various other distillation methods and summarize the results in Table~\ref{tbl:cifar100_transfer_more}. In general, CRD achieves the best transferring accuracy in 7 out of 10 settings.

\begin{table}[ht]

\setlength{\tabcolsep}{6pt}
\begin{center}
\begin{tabular}{lcccccccccc}
\toprule
\thead{Teacher \\ Student} & 

\multicolumn{2}{c}{\thead{WRN-40-2 \\ WRN-16-2}} &
\multicolumn{2}{c}{\thead{resnet56\\resnet20}} &
\multicolumn{2}{c}{\thead{resnet32x4\\resnet8x4}} &
\multicolumn{2}{c}{\thead{vgg13 \\ MobileNetV2}} &
\multicolumn{2}{c}{\thead{resnet32x4\\ShuffleNetV2}} \\

\midrule

dataset & STL & TI & STL & TI & STL & TI & STL & TI & STL & TI\\

\midrule

Vanilla & 69.7 & 33.7 & 69.1 & 30.4 & 71.8 & 36.7 & 64.2 & 29.1 & 65.1 & 28.9 \\
KD & 70.9 & 33.9 & 69.1 & 32.0 & 71.8 & 36.2 & 66.3 & 29.9 & 69.5 & 33.4 \\
FitNet & 70.3 & 33.5 & 68.1 & 29.5 & 73.5 & 40.2 & 66.9 & 31.4 & 71.7 & 36.2 \\
AT & 70.7 & 34.2 & 70.1 & 31.0 & 74.1 & 39.2 & 66.3 & 28.4 & 72.5 & 37.3 \\
SP & 71.3 & 34.1 & 68.0 & 30.2 & 72.9 & 37.6 & 67.8 & 31.6 & 68.8 & 33.4 \\
CC & 70.5 & 33.7 & 68.6 & 30.4 & 72.1 & 37.1 & 66.3 & 30.6 & 69.9 & 34.4 \\
VID & 71.0 & 34.6 & 69.9 & 32.4 & 73.9 & 39.6 & 66.1 & 31.1 & 71.4 & 36.5 \\
RKD & 71.6 & 35.1 & \underline{70.6} & 32.6 & 73.9 & 38.8 & 68.4 & 32.9 & 71.5 & 37.8 \\
PKT & 71.6 & 34.8 & 69.1 & 31.6 & 73.5 & 38.0 & 69.5 & 33.5 & 70.7 & 36.4 \\
AB & 70.8 & 32.2 & 68.5 & 29.7 & 73.4 & 37.6 & 67.6 & 31.0 & 71.3 & 36.2 \\
FT & \underline{71.7} & 34.7 & \textbf{70.9} & 33.5 & \textbf{75.2} & \textbf{40.4} & 68.9 & 32.5 & 72.9 & 38.7 \\
FSP & 69.6 & 33.6 & 68.1 & 30.4 & 72.4 & 37.1 & n/a & n/a & n/a & n/a \\
NST & 70.2 & 32.6 & 68.5 & 30.1 & 73.9 & 38.7 & 64.9 & 27.8 & 72.3 & 37.8 \\
\midrule
CRD & 71.6 & \textbf{35.6} & 70.2 & \textbf{34.3} & \underline{74.8} & \underline{40.2} & \textbf{71.6} & \textbf{35.7} & \textbf{73.5} & \textbf{40.1} \\
CRD+KD & \textbf{72.2} & \underline{35.5} & 70.5 & \underline{34.1} & 74.6 & 39.3 & \underline{70.7} & \underline{34.6} & \underline{73.1} & \underline{39.9} \\

\end{tabular}
\vspace{-5pt}
\caption{
\small{
We measure the transferability of the student network, by evaluating a linear classifier on top of its frozen representations on STL10 (abbreviated as ``STL'') and TinyImageNet (abbreviated as ``TI''). The best accuracy is bolded and the second best is underlined.
}}
\label{tbl:cifar100_transfer_more}
\end{center}
\vspace{-10pt}
\end{table}

\subsection{Deep Mutual Learning Setting}

In~\citep{zhang2018deep}, a Deep Mutual Learning setting was proposed where the teacher and student networks are trained simultaneously rather than sequentially. The benefit is that not only the student networks but also the teacher models will be improved. Here we investigate the possibility of incorporating different distillation objectives into this training framework. The results are summarizied in Table~\ref{tbl:mutual}. We observe that in general logit based distillation methods is better than non-logit based distillation methods in this setting, i.e., KD performs the best. We conjecture that during the mutual training phase, KD performs like an advanced label smoothing regularization and does not require the logit to be very accurate. But the feature-based distillation methods are hard to learn knowledge from the teacher model when it is handicapped. On the other hand, we notice that the combination of KD and CRD leads to better performance on the student side, as shown in the last row of Table~\ref{tbl:mutual}.

\begin{table}[ht]

\setlength{\tabcolsep}{3.5pt}
\begin{center}
\begin{tabular}{lccccccccc}
\toprule
\thead{Teacher \\ Student} & 
\thead{WRN-40-2 \\ WRN-16-2} & 
\thead{WRN-40-2 \\ WRN-40-1} & 
\thead{resnet56 \\ resnet20} & 
\thead{resnet110 \\ resnet20} & 
\thead{resnet110 \\ resnet32} & 
\thead{resnet32x4 \\ resnet8x4} & 
\thead{vgg13\\vgg8} &
\thead{ResNet50 \\ vgg8}\\
\midrule
Vanilla & 75.61 & 75.61 & 72.34 & 74.31 & 74.31 & 79.42 & 74.64 & 79.34 &\multirow{7}{*}{T}\\
KD & 77.89 & \textbf{77.42} & \textbf{74.51} & \textbf{75.95} & \textbf{76.70} & \textbf{79.90} & \textbf{76.53} & \textbf{79.92} \\
AT & 76.02 & 76.42 & 72.44 & 73.72 & 73.69 & 79.08 & 73.21 & 79.33 \\
SP & 76.36 & 76.08 & 73.11 & 73.58 & 73.81 & 79.50 & 75.75 & 73.34 \\
CC & 76.42 & 76.30 & 71.95 & 72.78 & 74.19 & 79.44 & 74.80 & 77.76 \\
CRD & 77.30 & 77.23 & 73.25 & 75.32 & 74.37 & 79.81 & 75.25 & 78.65 \\
CRD+KD & \textbf{78.01} & 77.39 & 73.86 & 75.31 & 75.53 & 79.36 & 77.23 & 79.04 \\
\midrule
Vanilla & 73.26 & 71.98 & 69.06 & 69.06 & 71.14 & 72.50 & 70.36 & 70.36 &\multirow{7}{*}{S}\\
KD & 74.98 & 73.66 & 70.85 & 70.71 & \textbf{73.24} & 74.77 & 73.39 & 74.00 \\
AT & 73.04 & 71.57 & 69.20 & 69.04 & 70.87 & 72.32 & 69.55 & 69.56 \\
SP & 72.82 & 71.47 & 69.45 & 69.31 & 70.75 & 72.63 & 70.57 & 70.66 \\
CC & 73.19 & 71.46 & 68.82 & 69.51 & 71.24 & 72.57 & 71.12 & 70.36 \\
CRD & 75.22 & 73.53 & 70.92 & 70.80 & 72.65 & 75.24 & 73.19 & 73.21 \\
CRD+KD & \textbf{75.89} & \textbf{74.12} & \textbf{70.90} & \textbf{71.60} & 73.07 & 75.34 & \textbf{74.08} & \textbf{74.22} \\

\end{tabular}
\vspace{-5pt}
\caption{
\small{
Test \emph{accuracy} (\%) of student and teacher networks on CIFAR100 with the Deep Mutual Training~\cite{zhang2018deep} setting, where the teacher and student networks are trained simultaneously rather than sequentially. We use ``T'' and ``S'' to denote the teacher and student models, respectively.
}}
\label{tbl:mutual}
\end{center}
\vspace{-10pt}
\end{table}

\subsection{Standard Deviation for Results on CIFAR-100 Benchmark}

The standard deviation over multiple runs on CIFAR-100 benchmark is provided in Table~\ref{tbl:cifar100_same_std} for student and teacher models of the same architectural type, and in Table~\ref{tbl:cifar100_diff_std} for student and teacher models of different architectural types.

\newpage
\begin{table}[ht]
\setlength{\tabcolsep}{2.5pt}
\begin{center}
\begin{small}
\begin{tabular}{lccccccc}
\toprule
\thead{Teacher \\ Student} & \thead{WRN-40-2 \\ WRN-16-2} & \thead{WRN-40-2 \\ WRN-40-1} & \thead{resnet56\\resnet20} & \thead{resnet110\\resnet20} & \thead{resnet110\\resnet32} & \thead{resnet32x4\\resnet8x4} & \thead{vgg13\\vgg8}\\
\midrule
Teacher & 75.61 & 75.61 & 72.34 & 74.31 & 74.31 & 79.42 & 74.64 \\
Student & 73.26 & 71.98 & 69.06 & 69.06 & 71.14 & 72.50 & 70.36 \\
\midrule

KD & 74.92 $\pm$ 0.28& 73.54 $\pm$ 0.20& 70.66 $\pm$ 0.24& 70.67 $\pm$ 0.27& 73.08 $\pm$ 0.18& 73.33 $\pm$ 0.25& 72.98 $\pm$ 0.19\\
FitNet & 73.58 $\pm$ 0.32& 72.24 $\pm$ 0.24& 69.21 $\pm$ 0.36& 68.99 $\pm$ 0.27& 71.06 $\pm$ 0.13& 73.50 $\pm$ 0.28& 71.02 $\pm$ 0.31\\
AT & 74.08 $\pm$ 0.25& 72.77 $\pm$ 0.10& 70.55 $\pm$ 0.27& 70.22 $\pm$ 0.16& 72.31 $\pm$ 0.08& 73.44 $\pm$ 0.19& 71.43 $\pm$ 0.09\\
SP & 73.83 $\pm$ 0.12& 72.43 $\pm$ 0.27& 69.67 $\pm$ 0.20& 70.04 $\pm$ 0.21& 72.69 $\pm$ 0.41& 72.94 $\pm$ 0.23& 72.68 $\pm$ 0.19\\
CC & 73.56 $\pm$ 0.26& 72.21 $\pm$ 0.25& 69.63 $\pm$ 0.32& 69.48 $\pm$ 0.19& 71.48 $\pm$ 0.21& 72.97 $\pm$ 0.17& 70.71 $\pm$ 0.24\\
VID & 74.11 $\pm$ 0.24& 73.30 $\pm$ 0.13& 70.38 $\pm$ 0.14& 70.16 $\pm$ 0.39& 72.61 $\pm$ 0.28& 73.09 $\pm$ 0.21& 71.23 $\pm$ 0.06\\
RKD & 73.35 $\pm$ 0.09& 72.22 $\pm$ 0.20& 69.61 $\pm$ 0.06& 69.25 $\pm$ 0.05& 71.82 $\pm$ 0.34& 71.90 $\pm$ 0.11& 71.48 $\pm$ 0.05\\
PKT & 74.54 $\pm$ 0.04& 73.45 $\pm$ 0.19& 70.34 $\pm$ 0.04& 70.25 $\pm$ 0.04& 72.61 $\pm$ 0.17& 73.64 $\pm$ 0.18& 72.88 $\pm$ 0.09\\
AB & 72.50 $\pm$ 0.26& 72.38 $\pm$ 0.31& 69.47 $\pm$ 0.09& 69.53 $\pm$ 0.16& 70.98 $\pm$ 0.39& 73.17 $\pm$ 0.31& 70.94 $\pm$ 0.18\\
FT & 73.25 $\pm$ 0.20& 71.59 $\pm$ 0.15& 69.84 $\pm$ 0.12& 70.22 $\pm$ 0.10& 72.37 $\pm$ 0.31& 72.86 $\pm$ 0.12& 70.58 $\pm$ 0.08\\
FSP & 72.91 $\pm$ 0.24& n/a& 69.95 $\pm$ 0.21& 70.11 $\pm$ 0.05& 71.89 $\pm$ 0.11& 72.62 $\pm$ 0.13& 70.23 $\pm$ 0.23\\
NST & 73.68 $\pm$ 0.11& 72.24 $\pm$ 0.22& 69.60 $\pm$ 0.13& 69.53 $\pm$ 0.15& 71.96 $\pm$ 0.07& 73.30 $\pm$ 0.28& 71.53 $\pm$ 0.13\\
CRD & 75.48 $\pm$ 0.09& 74.14 $\pm$ 0.22& 71.16 $\pm$ 0.17& 71.46 $\pm$ 0.09& 73.48 $\pm$ 0.13& 75.51 $\pm$ 0.18& 73.94 $\pm$ 0.22\\

\midrule
CRD+KD & 75.64 $\pm$ 0.21& 74.38 $\pm$ 0.11& 71.63 $\pm$ 0.15& 71.56 $\pm$ 0.16& 73.75 $\pm$ 0.24& 75.46 $\pm$ 0.25& 74.29 $\pm$ 0.12\\

\end{tabular}
\end{small}
\vspace{-5pt}
\caption{
\small{
Test \emph{accuracy} (\%) of student networks on CIFAR100 of a number of distillation methods (ours is CRD). Standard deviation is provided.
}}
\label{tbl:cifar100_same_std}
\end{center}
\vspace{-10pt}
\end{table}
\begin{table}[ht]

\setlength{\tabcolsep}{4.5pt}
\begin{center}
\begin{small}
\begin{tabular}{lcccccc}
\toprule
\thead{Teacher \\ Student} & \thead{vgg13 \\ MobileNetV2} & \thead{ResNet50 \\ MobileNetV2} & \thead{ResNet50 \\ vgg8} & \thead{resnet32x4\\ShuffleNetV1} & \thead{resnet32x4\\ShuffleNetV2} & \thead{WRN-40-2\\ShuffleNetV1}\\
\midrule
Teacher & 74.64	& 79.34 & 79.34	& 79.42	& 79.42	& 75.61 \\
Student & 64.6 & 64.6 & 70.36 & 70.5 & 71.82 & 70.5 \\
\midrule

KD & 67.37 $\pm$ 0.32& 67.35 $\pm$ 0.32& 73.81 $\pm$ 0.13& 74.07 $\pm$ 0.19& 74.45 $\pm$ 0.27& 74.83 $\pm$ 0.17\\
FitNet & 64.14 $\pm$ 0.50& 63.16 $\pm$ 0.47& 70.69 $\pm$ 0.22& 73.59 $\pm$ 0.15& 73.54 $\pm$ 0.22& 73.73 $\pm$ 0.32\\
AT & 59.40 $\pm$ 0.20& 58.58 $\pm$ 0.54& 71.84 $\pm$ 0.28& 71.73 $\pm$ 0.31& 72.73 $\pm$ 0.09& 73.32 $\pm$ 0.35\\
SP & 66.30 $\pm$ 0.38& 68.08 $\pm$ 0.38& 73.34 $\pm$ 0.34& 73.48 $\pm$ 0.42& 74.56 $\pm$ 0.22& 74.52 $\pm$ 0.24\\
CC & 64.86 $\pm$ 0.25& 65.43 $\pm$ 0.15& 70.25 $\pm$ 0.12& 71.14 $\pm$ 0.06& 71.29 $\pm$ 0.38& 71.38 $\pm$ 0.25\\
VID & 65.56 $\pm$ 0.42& 67.57 $\pm$ 0.28& 70.30 $\pm$ 0.31& 73.38 $\pm$ 0.09& 73.40 $\pm$ 0.17& 73.61 $\pm$ 0.12\\
RKD & 64.52 $\pm$ 0.45& 64.43 $\pm$ 0.42& 71.50 $\pm$ 0.07& 72.28 $\pm$ 0.39& 73.21 $\pm$ 0.28& 72.21 $\pm$ 0.16\\
PKT & 67.13 $\pm$ 0.30& 66.52 $\pm$ 0.33& 73.01 $\pm$ 0.14& 74.10 $\pm$ 0.25& 74.69 $\pm$ 0.34& 73.89 $\pm$ 0.16\\
AB & 66.06 $\pm$ 0.48& 67.20 $\pm$ 0.37& 70.65 $\pm$ 0.09& 73.55 $\pm$ 0.31& 74.31 $\pm$ 0.11& 73.34 $\pm$ 0.09\\
FT & 61.78 $\pm$ 0.33& 60.99 $\pm$ 0.37& 70.29 $\pm$ 0.19& 71.75 $\pm$ 0.20& 72.50 $\pm$ 0.15& 72.03 $\pm$ 0.16\\
NST & 58.16 $\pm$ 0.26& 64.96 $\pm$ 0.44& 71.28 $\pm$ 0.13& 74.12 $\pm$ 0.19& 74.68 $\pm$ 0.26& 74.89 $\pm$ 0.25\\
CRD & 69.73 $\pm$ 0.42& 69.11 $\pm$ 0.28& 74.30 $\pm$ 0.14& 75.11 $\pm$ 0.32& 75.65 $\pm$ 0.10& 76.05 $\pm$ 0.14\\

\midrule

CRD+KD & 69.94 $\pm$ 0.05& 69.54 $\pm$ 0.39& 74.58 $\pm$ 0.27& 75.12 $\pm$ 0.35& 76.05 $\pm$ 0.09& 76.27 $\pm$ 0.29\\

\end{tabular}
\end{small}
\vspace{-5pt}
\caption{
\small{
Top-1 test \emph{accuracy} (\%) of student networks on CIFAR100 of a number of distillation methods (ours is CRD) for transfer across very different teacher and student architectures. Standard deviation is provided.
}}
\label{tbl:cifar100_diff_std}
\end{center}
\vspace{-10pt}
\end{table}

\end{document}


\maketitle

\section{Hyper-parameters}
\vspace{-5pt}
We optimize the hyper-parameters for all objectives on CIFAR-100 with teacher WRN-40-2 and WRN-16-2 on a validation set splitted out from the $50000$ training images. Then we apply these hyper-parameters to other experiments.

\textbf{Knowledge Distillation (KD).} For this objective, we search $\alpha$ in $\{0.1,0.3,0.5,0.7,0.9,0.95,0.99\}$ and $\rho$ in $\{2, 4, 7, 10, 15, 20, 30\}$, finding that $\alpha=0.9$ and $\rho=4$ in general works the best. This agrees with the setting used in~\cite{zagoruyko2016paying}. But in ImageNet, we find using a weight of $0.5$ for hard label cross-entropy $H(y, y^S)$ and a weight of $0.9$ for soft label cross-entropy $H(\sigma(z^T/\rho), \sigma(z^S/\rho))$ performs better. Therefore, the accuracy of KD on ImageNet shown in Table 2 is reported with this setting.

\textbf{Attention Transfer (AT).} For this objective, we search $\beta$ in $\{100, 500, 800, 1000, 2000\}$ and use $1000$ across all experiments.

\textbf{Regression (FitNet).} For this objective, we search $\beta$ in $\{20, 50, 100, 200, 500\}$ and use $100$ across all experiments.

\textbf{Contrastive Knowledge Distillation (CKD).} For this objective, we search $\beta$ in $\{0.2, 0.5, 0.8, 1.2, 2, 3, 5\}$ and use $0.8$ across all experiments. 
\vspace{-5pt}

\section{Training and validation curves on ImageNet}
\vspace{-5pt}
We plot the training and validation curves of different distillation objectives on ImageNet dataset, as shown in Fig.~\ref{fig:imagenet}. Our CKD achieves better resutls on the validation set.
\begin{figure}[t]
\centering
\includegraphics[width=\linewidth]{fig/imagenet.pdf}
\vspace{-10pt}
\caption{
\small{
The training and validation accuracies on ImageNet when we distill knowledge from ResNet-34 to ResNet-18. Objectives include KD, AT and our CRD.
}
}
\label{fig:imagenet}
\end{figure}

\section{Visualization of the correlation discrepancy}
\begin{figure}[t]
\centering
\vspace{-20pt}
\includegraphics[width=\linewidth]{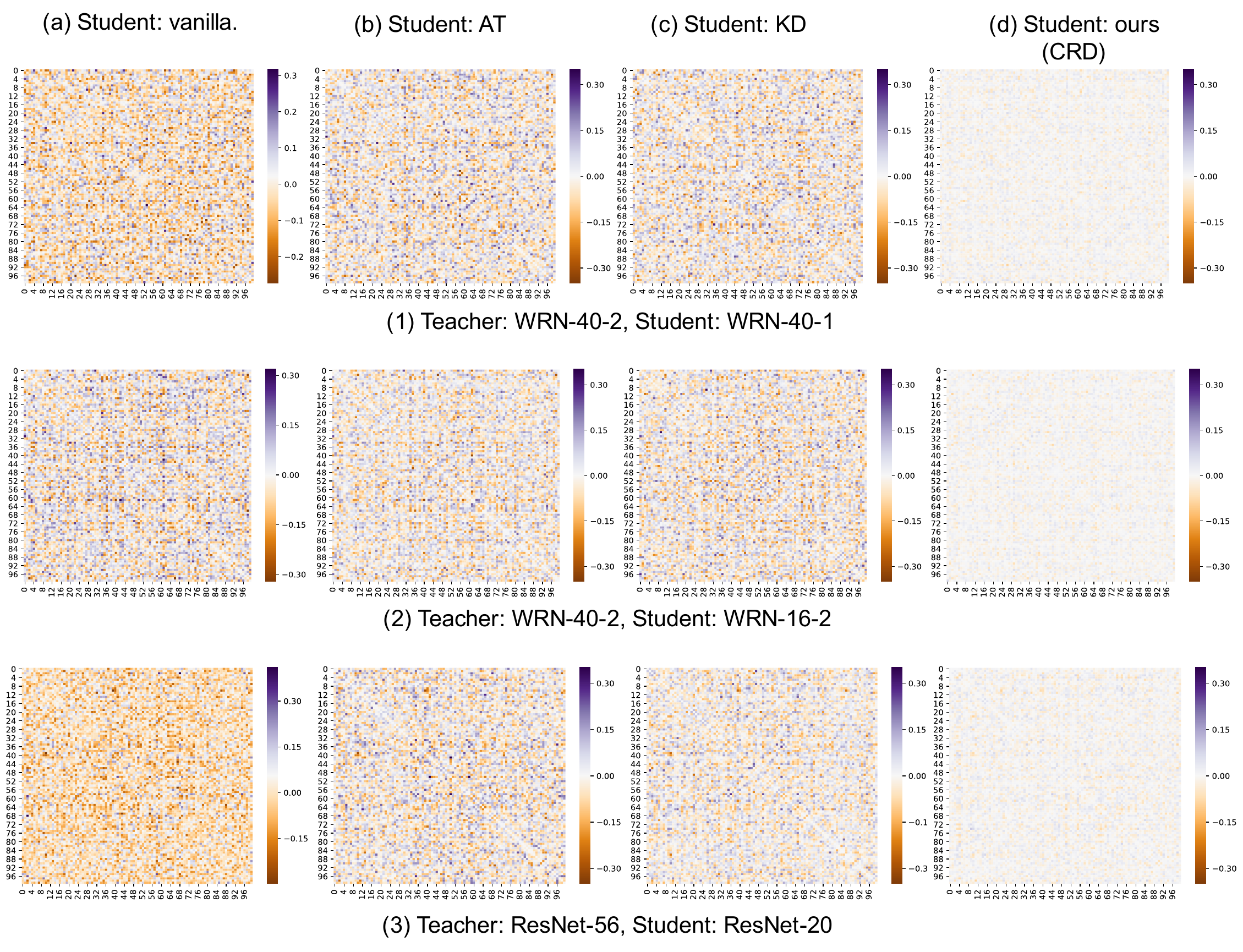}
\vspace{-10pt}
\caption{\small{The correlations between class logits output by the teacher network show the  ``dark knowledge" \cite{hinton2015distilling} that must be transferred to a student networks. A student network that captures these correlations tends to perform better at the task. We visualize here the difference between correlation matrices of the student and teacher at the logits, for different student networks on a Cifar100 knowledge distillation task: (a) A student trained without distillation; (b) A student distilled by attention transfer~\cite{zagoruyko2016paying} (c) A student distilled by KL divergence \cite{hinton2015distilling}; (d) A student distilled by our contrastive objective. Our objective greatly improves the structured knowledge (correlations) in the output units. 
}}
\label{fig:correlation_more}
\end{figure}

\vspace{-5pt}
We visualize the correlation discrepancy for different distillation objectives across various combinations of student and teacher networks. As shown in Fig.~\ref{fig:correlation_more}, our contrastive distillation objective siginificantly outperforms other objectives, in terms of minimizing the correlation discrepancy between student and teacher networks. The normalized correlation coefficients are computed at the logit layer.

\bibliographystyle{plain}
\bibliography{./refs.bib}